\documentclass[lettersize,journal]{IEEEtran}
\usepackage{amsmath,amsfonts}
\usepackage{algorithmic}
\usepackage{algorithm}
\usepackage{array}
\usepackage{textcomp}
\usepackage{float}
\usepackage{stfloats}
\usepackage{url}
\usepackage{verbatim}
\usepackage{subfig}
\usepackage{adjustbox}
\usepackage{booktabs}
\usepackage{graphicx}
\usepackage{multirow}
\usepackage{cite}
\usepackage{aliascnt}
\usepackage{color} 
\usepackage{makecell}
\usepackage[switch]{lineno}
\usepackage{soul}
\soulregister\cite7
\soulregister\ref7 
\usepackage{color, xcolor}
\definecolor{myblue}{RGB}{66,133,244}
\definecolor{mygreen}{RGB}{51,168,83}
\definecolor{myyellow}{RGB}{251,188,3}
\definecolor{myred}{RGB}{234,67,53}
\definecolor{mygrey}{RGB}{95,99,104}
\definecolor{mypup}{RGB}{153,0,204}

\begin{document}

\title{Adaptive Domain Generalization via Online Disagreement Minimization}

\author{Xin Zhang and Ying-Cong Chen,~\IEEEmembership{Member,~IEEE}
\thanks{Corresponding author: Ying-Cong Chen.}
\thanks{Xin Zhang is with the AI Thrust, Information Hub, Hong Kong University of Science and Technology (Guangzhou), Guangzhou, China (e-mail: devinxzhang@ust.hk).}
\thanks{Ying-Cong Chen is with the AI Thrust, Information Hub, Hong Kong University of Science and Technology (Guangzhou), Guangzhou, China, also with the Department of Computer Science and Engineering, Hong Kong University of Science and Technology, and also with the HKUST (Guangzhou) - SmartMore Joint Lab, Guangzhou, China (e-mail: yingcongchen@ust.hk).}
\thanks{This work is supported by National Natural Science Foundation of China (No. 62206068).}
}

\markboth{} 
{Shell \MakeLowercase{\textit{et al.}}: A Sample Article Using IEEEtran.cls for IEEE Journals}


\maketitle

\begin{abstract}
Deep neural networks suffer from significant performance deterioration when there exists distribution shift between deployment and training. Domain Generalization (DG) aims to safely transfer a model to unseen target domains by only relying on a set of source domains. 
Although various DG approaches have been proposed, a recent study named DomainBed~\cite{gulrajani2020search}, reveals that most of them do not beat simple empirical risk minimization (ERM). To this end, we propose a general framework that is orthogonal to existing DG algorithms and could improve their performance consistently. 
Unlike previous DG works that stake on a \textsl{static} source model to be hopefully a universal one, our proposed AdaODM \textsl{adaptively} modifies the source model at test time for different target domains. 
Specifically, we create multiple domain-specific classifiers upon a shared domain-generic feature extractor. 
The feature extractor and classifiers are trained in an adversarial way, where the feature extractor embeds the input samples into a domain-invariant space, and the multiple classifiers capture the distinct decision boundaries that each of them relates to a specific source domain. 
During testing, distribution differences between target and source domains could be effectively measured by leveraging prediction disagreement among source classifiers. 
By fine-tuning source models to minimize the disagreement at test time, target-domain features are well aligned to the invariant feature space. 
We verify AdaODM on two popular DG methods, namely ERM and CORAL, and four DG benchmarks, namely VLCS, PACS, OfficeHome, and TerraIncognita. 
The results show AdaODM stably improves the generalization capacity on unseen domains and achieves state-of-the-art performance.
\end{abstract}

\begin{IEEEkeywords}
Domain shift, domain generalization, online adaptation, consistency regularization.
\end{IEEEkeywords}

\section{Introduction}
\IEEEPARstart{D}{eep} neural networks work best when training and testing data are sampled from the same distribution~\cite{vapnik1999overview}. However, this is not always achievable in most real-world scenarios. 
For instance, medical images acquired by different scanners, vendors, and operators may show visible differences. Autonomous driving system must handle variations in light and weather conditions. 
 
Unfortunately, the distribution shift is notorious to harm the model's generalization ability on unseen data. A general solution to tackle this problem is Domain Adaptation (DA)~\cite{wilson2020survey,csurka2017domain}. 
DA assumes (unlabeled) samples of the target domain are available during training and alleviates domain shift by minimizing the discrepancy between source and target domain. 
Despite its promising performance, domain adaptation is not applicable when target domain is inaccessible during training. 
Besides, it also requires retraining the models when target domain changes (e.g., the system is moved to a new environment), which may cause large training costs. 
Domain Generalization (DG)~\cite{peng2019moment,ganin2015unsupervised,tzeng2017adversarial} relieves the constraint. 
It aims to learn a model that is generalizable to unseen target domains solely on multiple source domains, without access to any samples from target domain. 
Despite the ambitious goal, a recent study on large-scale benchmarks~\cite{gulrajani2020search} shows that most existing DG methods actually do not bring significant benefits compared with the naive Empirical Risk Minimization (ERM). 
This might imply that it is difficult to handle various unknown domain samples merely with a generalized source model trained via the DG protocol. 

\begin{figure}[t]
    \centering
    \includegraphics[width=9cm]{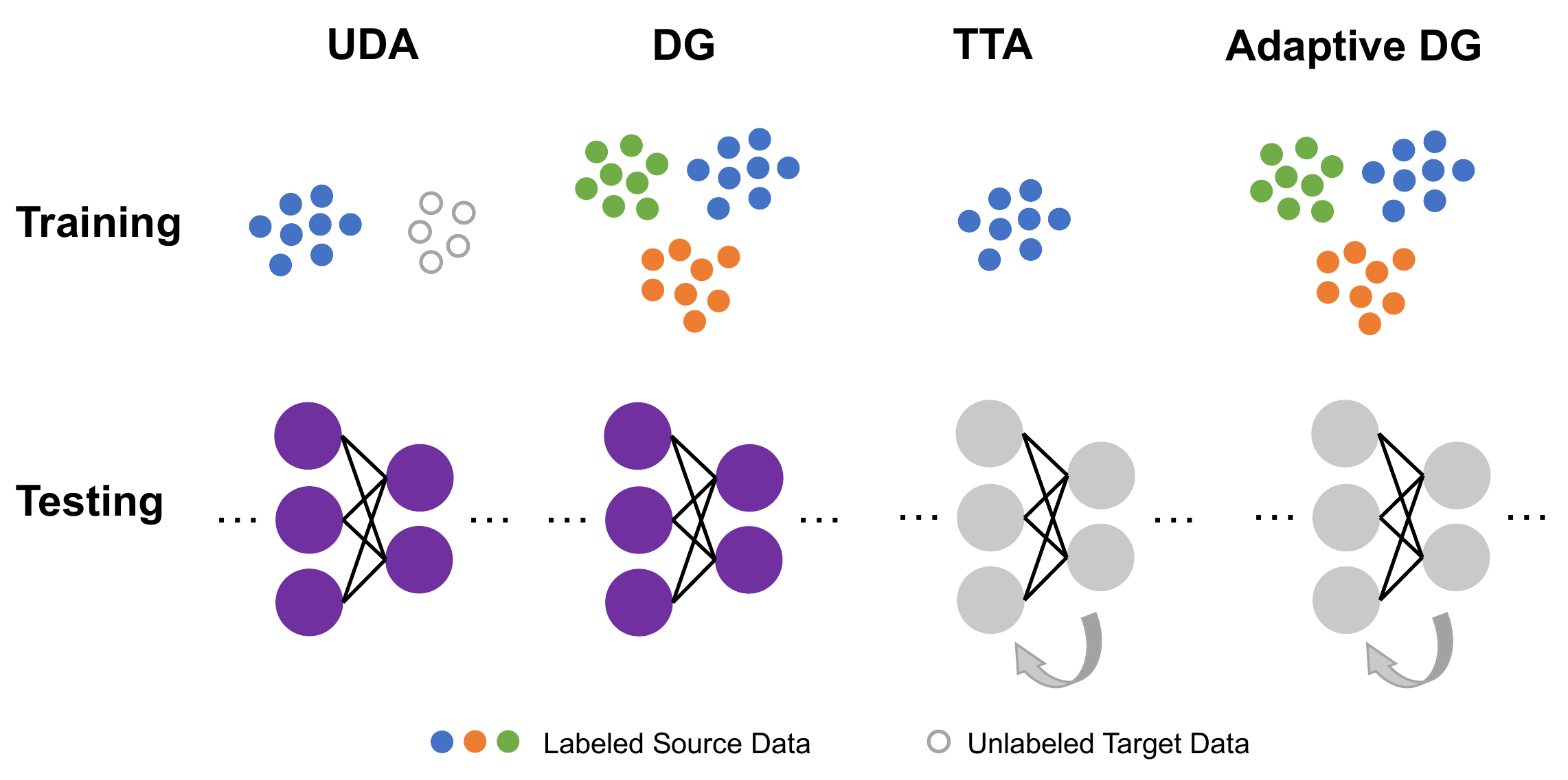}
    \caption{Comparison of UDA, DG, TTA, and Adaptive DG settings. UDA assumes access to unlabeled target data during training. DG learns from multiple source domains to improve models' generalization ability. TTA adapts a source model to target domains via test-time online adaptation. Adaptive DG takes advantages of both DG and TTA to better leverage domain information for adaptively adjusting source models for new domains. }
    \label{fig:tta}
\end{figure}

Therefore, we are particularly interested in how to facilitate DG to further utilize target domain information. 
Recently, the Test-Time Adaptation (TTA) paradigm has become popular in resolving the domain shift problem which updates the source model to mitigate the domain gap at test time. 
TTA is based on the fact that during testing, the model can always access query samples of the target domain which provides essential target domain information. 
We introduce the idea of TTA to the DG field to adaptively adjust the DG model based on the target domain samples during testing (See Fig. \ref{fig:tta} for comparison).

\begin{figure*}[t]
    \centering
    \includegraphics[width=15cm]{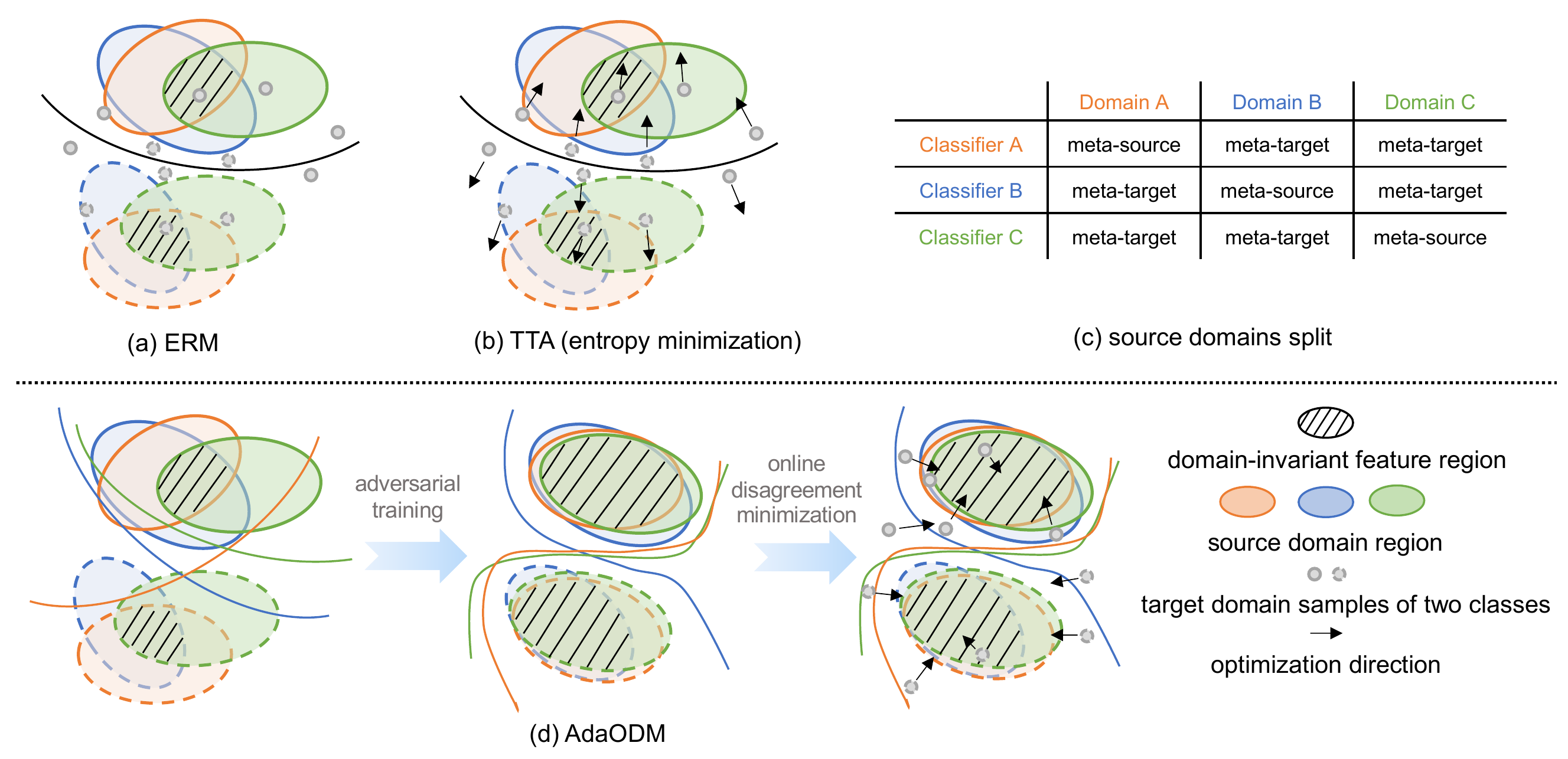}
    \caption{Comparison of ERM, Entropy Minimization (EM)-based Test-Time Adaptation (TTA) method, and the proposed AdaODM. Here we take for instance a two-class classification task with three source domains. Solid and dashed circles represent two classes. Colors represent various domains. (a) shows the vanilla ERM which performs well on three training sources while predicting wrongly on some target samples (dashed points above the line and solid points below the line). In (b), the EM-based TTA aims to make more confident predictions on target domains by pushing target features away from the decision boundary. This could reduce domain gaps to some extent. However, those falsely predicted samples are pushed in wrong directions, which harms adaptation performance. AdaODM constructs domain-specific classifiers as (d). During training, AdaODM promotes each classifier's generalization capacity by encouraging prediction consistency on unseen domains (``meta-target'' domains in (c)). This is done through adversarial training where the feature extractor minimizes the prediction disagreement while classifiers maximize it. Consequently, we achieve domain-invariant feature space encompassed by class boundaries. At test time, classifiers are frozen. We optimize the feature extractor to minimize the prediction disagreement on target samples, which aligns target features to domain-invariant regions.}
    \label{fig:main}
\end{figure*}

Existing TTA works include updating the batch-norm statistics based on testing data~\cite{hu2021mixnorm,schneider2020removing},  designing unsupervised loss functions (e.g., entropy minimization) to update the feature extractor~\cite{liang2020we,wang2020tent}, or adjusting the classifier based on testing samples~\cite{iwasawa2021test}. 
Although effective, these methods do not take domain differences into consideration. Domain differences provide valuable information to find the \textsl{invariance} shared by all domains.
Without leveraging this information, the model may fail in handling target domains with large domain shift~\cite{luo2008transfer,peng2019moment}. 
Moreover, losses like entropy minimization can only enforce target features away from decision boundaries whose optimization direction might be less reliable (see Fig. \ref{fig:main}(b)). 
Our adaptive domain generalization aims to better leverage both source and target domains information to measure the distribution differences, and models are optimized towards aligning target domain features to be invariant to domain changes. 

Motivated by the above considerations, we propose a novel test-time domain generalization framework named \textbf{AdaODM} which \textbf{Ada}ptively adjusts the source DG model based on the target domain data via \textbf{O}nline \textbf{D}isagreement \textbf{M}inimization. 
Our idea is to introduce a set of domain-aware classifiers, each of which is trained with data from one single domain. As such, these classifiers may leverage both domain-generic and domain-specific features. During testing, these classifiers may generate different prediction scores because of the interference of domain-specific features. Therefore, by minimizing the prediction disagreement, we can reduce the influence of fragile features that are specific for each domain, leading to better performance of out-of-domain samples. Note that such optimization is model-agnostic and can be performed during testing. This allows us to facilitate existing DG methods to adapt source models for different target domains.

Although the idea looks neat, training these classifiers is challenging. 
This is because each classifier is trained only on a subset of total source domains, and thus they are more likely to exhibit higher variance. 
As such, simply minimizing the prediction disagreement may not always lead to accurate predictions, as it can cause the classifiers to agree on incorrect classes. To solve this problem, we propose a novel adversarial training scheme to improve each classifier's generalization ability on other domains during training. 
Specifically, we optimize the feature extractor to minimize the prediction disagreement among source classifiers, and optimize the classifiers to maximize it. 
Consequently, the feature extractor becomes more invariant to domain changes, while the classifiers are more sensitive to distribution differences among different domains. 
Such a process can also be seen as achieving domain-invariant feature space encompassed by domain-specific class decision boundaries (see Fig. \ref{fig:main}(d)). 
During inference, source classifiers are frozen and the feature extractor is adjusted to reduce the prediction disagreement among classifiers. 
The main contributions of this paper can be summarized as follows. 
\begin{itemize}
    \item We propose a novel adaptive domain generalization framework, which could dynamically adjust source DG models to target domains at test time via online prediction disagreement minimization. 
    \item We capture domain information with domain-specific classifiers and present a novel adversarial training scheme to learn domain-invariant features. 
    \item We evaluate our framework on two DG methods and four DG benchmarks and achieve state-of-the-art performance. 
\end{itemize}

\section{Related Work}

\subsection{Domain Adaptation and Generalization}
Unsupervised Domain Adaptation (UDA)~\cite{wilson2020survey,csurka2017domain} jointly optimizes the model on labeled source domain images and unlabeled target domain images to mitigate the domain gap. 
Feature alignment-based methods focus on matching the distributions of feature activations between source and target data. 
This is done typically in two manners: (1) moment matching schemes~\cite{long2017deep,gretton2012kernel,sun2016deep} aim to explicitly minimize some specific statistics distances between the two domains; (2) adversarial learning-based approaches~\cite{ganin2016domain,saito2018maximum,long2018conditional} leverage a domain discriminator to encourage the domain confusion. 
In particular, MCD~\cite{saito2018maximum} utilizes classifier discrepancy as a measure to optimize both the feature extractor and the classifier via adversarial training. 
Other works also explore the distribution alignment in data space~\cite{hoffman2018cycada}. 
UDA has to access unlabeled target domain data and retrain the model for new target domains, which makes it impractical for real-world applications. 
This is improved by recent ``source-free'' DA methods~\cite{liang2020we,li2020model,kundu2020universal} which adapt the source model solely utilizing unlabeled target data. 
However, these methods generally require access to all target data during adaptation and optimize source models for multiple epochs, making them impractical for online adaptation. Instead, AdaODM aims to bridge the domain gap while performing inference, in which case the test samples pass through the model only once without being revisited.

The goal of domain generalization is to learn a generalized model on multiple source domains without relying on any target domain information. 
Like UDA, a common pervasive theme in domain generalization aims to capture not only the task-discriminative but the \textsl{domain-invariant} representations~\cite{li2018deep,li2018domain,senerdomain,dingdomain}.
The underlying principle is that the mapping relationship from input to output is consistent and invariant to environment changes, DG solutions aim to estimate various types of invariance from data~\cite{gulrajani2020search}. 
Meta-learning-based DG methods~\cite{li2018learning,dou2019domain,balaji2018metareg} simulate domain shift by splitting the source domains into meta-train and meta-test and learn transferable weight representations via episodic training. 
Invariant risks minimization~\cite{arjovsky2019invariant} regularizes ERM to explore invariant predictors across all domains. 
Other approaches consider various domain augmentation strategies~\cite{wang2020heterogeneous,xu2020adversarial,yan2020improve,zhou2020domain,jeon2021feature,wang2021feature} to facilitate the inter-domain constraint. 
Although a large number of studies have been developed to address the difficult domain generalization issue, a comprehensive benchmark comparison conducted by Gulrajani and Lopez-Paz~\cite{gulrajani2020search} shows that ERM is still a strong baseline method compared with most state-of-the-art approaches. 

\subsection{Test-Time Training/Adaptation}
Recently, adapting models solely on target data at test time has become appealing since test-time adaptation can leverage target data to adapt the source model online~\cite{wang2020tent,liang2020we,sun2020test,yang2022domain}. The distribution difference is estimated and adapted while a model makes the inference. 
Batch normalization statistics have shown to be essential to model robustness~\cite{schneider2020removing}. 
Some studies propose to replace the batch-norm statistics estimated on the training set with those on the test samples~\cite{schneider2020removing,hu2021mixnorm}. 
Tent~\cite{wang2020tent} further adapts affine parameters of batch-norm layers by minimizing the prediction entropy. 
Self-supervision is also widely used for test-time feature alignment~\cite{sun2020test,liang2020we,liu2021ttt++}. 
Test-time training~\cite{sun2020test,liu2021ttt++} jointly trains an auxiliary self-supervised task along with the main task during training, and optimizes the model with only the self-supervised task at test time. 
SHOT~\cite{liang2020we} updates the feature extractor by both information maximization and self-supervised pseudo-labeling~\cite{lee2013pseudo}. 
MEMO~\cite{zhang2021memo} extends Tent to enhance TTA's reliability by estimating prediction entropy over multiple augmented views of test examples. 
However, these methods do not consider the domain information among source domains for test-time adaptation. 

\subsection{Consistency Regularization}
Consistency regularization is widely used for semi-supervised~\cite{miyato2018virtual,xie2020unsupervised} and unsupervised learning~\cite{zhang2019consistency,sinha2021consistency}. 
Consistency regularization for semi-supervised learning regularizes model predictions to be invariant to semantics-preserving changes to images, such as rotation, crop, Gaussian noise, or adversarial attacking, so as to improve classification of unlabeled images. 
Mean-teacher~\cite{tarvainen2017mean} designs a consistency cost between predictions of the student and the teacher models, resulting in a better learning of the student model. 
Studies also exploit consistency regularizers to stabilize training of Generative Adversarial Networks (GANs)~\cite{zhang2019consistency} and solve the mismatch of latent representations learned by Variational Auto-Encoders (VAEs)~\cite{sinha2021consistency}. 
Recently, consistency regularization is also utilized in UDA or DG for feature alignment~\cite{french2017self,wu2020dual,xu2021fourier,saito2018maximum} or style transfer~\cite{hoffman2018cycada}. Inspired by these works, we design a disagreement score among predictions of source classifiers to further minimize distribution discrepancy at test time.

\section{Method}
Our method allows for adjusting the model parameters to adaptively alleviate the domain gaps during testing time. 
Specifically, our model consists of a domain-generic feature extractor and multiple domain-specific classifiers. 
During training, the feature extractor is trained to minimize the prediction disagreement, while the classifiers are trained to maximize it. 
This draws source domains closer and encourages a domain-invariant feature space revealed by the common nature of distinct classifiers. 
At test time, the classifiers are frozen and we optimize the feature extractor to reduce the prediction disagreement. 
This aligns query samples from target domains to the invariant feature space. We illustrate the proposed approach in Fig. \ref{fig:main}. 
In the following, we first give the basic formulation of domain generalization, then detail the prediction disagreement-based adversarial training. 
Finally, we will introduce how to adapt the trained source model to target domains via online disagreement minimization. It is worth noting that our algorithm is a general framework which could be implemented on most existing DG methods. We show case on vanilla Empirical Risk Minimization (ERM). 

\subsection{Problem Setup}
Without loss of generality, we focus on the image classification task which maps inputs $x\in\mathcal{X}$ to the corresponding output label $y\in\mathcal{Y}$ with model $f$, $f: \mathcal{X} \xrightarrow{} \mathcal{Y}$. The model $f$ can be further decomposed as a feature extractor $g$ followed by a classifier $h$, $f=h\circ g$.
Consider the problem of domain generalization, we assume the accessibility to training datasets $D^{d}=\{{(x^{d}_{i}, y^{d}_{i})}\}^{n_{d}}_{i=1}$ collected from multiple source domains $d\in \{1,...,d_{tr}\}$, where $d$, $n_{d}$ and $d_{tr}$ denote domain index, number of samples in domain $d$, and number of domains, respectively. 
Within each dataset $D^{d}$, samples are identically and independently distributed (i.i.d.) from the same joint probability density function $P(X^{d},Y^{d})$, where $X^{d} \in \mathcal{X}$ and $Y^{d} \in \mathcal{Y}$. 
Domain generalization aims to learn a generalized model $f$ on all source domain datasets $D^{d}$ to perform well on a new target domain $D^{d_{te}}$ where $P(X^{d_{te}},Y^{d_{te}}) \neq P(X^{d},Y^{d})$ for all $d\in\{1,...,d_{tr}\}$. 
Note that no prior information of target distribution $P(X^{d_{te}},Y^{d_{te}})$ is available during training, previous studies generally try to learn a generalized feature extractor $g$ to map samples from all domains into a domain-invariant feature space $\hat{\mathcal{S}}$, $g: x^{d} \xrightarrow{} \hat{S}$ for all $d \in \{1,...,d_{tr},d_{te}\}$. 
The adaptive domain generalization further adapts the source model $f^{tr}$ into a target one $f^{te}$ with bunches of non-revisitable target samples, during which source data are not accessible.

\subsection{Disagreement Score}
The goal of domain generalization is to achieve consistent mapping from the input $\mathcal{X}$ to the output label $\mathcal{Y}$ regardless of the distribution shift. 
An essential route is to learn a shared feature space $\hat{\mathcal{S}}$ for data from all domains. We consider the role of class boundaries, which are neglected by most existing DG methods, to capture source domain characteristics and achieve the \textsl{invariance} of feature space $\hat{\mathcal{S}}$ for both source and target domains. 
Specifically, we propose to learn a set of classifiers $\{h^{1}, ..., h^{d_{tr}}\}$ specific to each source domain $\{1,...,d_{tr}\}$ on top of a shared feature extractor, respectively. 
Suppose each domain-specific classifier $h^{d}$ could predict in-domain samples $x^{d}$ (we omit \textsl{class} notion for clarity) into correct categories well. 
With information from different domains, these domain-specific classifiers will be semantically related but show various decision boundaries. 
Intuitively, when an incoming target sample $x^{d_{te}}$ is classified consistently by all source classifiers, this indicates that the features of the sample are likely to be located in the shared region of $\hat{\mathcal{S}}$.
If the sample is far from the support of $\hat{\mathcal{S}}$, classifiers are likely to make inconsistent predictions because $x$ can be out of some classifiers' decision boundaries. As shown in the left of Fig. \ref{fig:main}(d), each domain-specific classifier does not perfectly classify samples in the other two source domains. 

Thus, we define a Disagreement Score ($\mathcal{DS}$) as the distance of prediction probabilities of any two source classifiers on target samples. Theoretically, there are many choices for computing the distance. In section \ref{sec:DS}, we show that $L_1$-norm works better than $L_2$-norm and $KL$ Divergence. We calculate $\mathcal{DS}$ as the sum of the $L_1$-norm of the difference of softmax outputs of any two source classifiers on predicting target samples:  
\begin{equation}
    \mathcal{DS} = \sum_{\substack{j,k \in \{1,...,d_{tr}\} \\j\neq k}}\frac{1}{n_{te}}\sum_{i=1}^{n_{te}}||s(h^{j}(g(x^{te}_{i})))-s(h^{k}(g(x^{te}_{i})))||_1.
    \label{ds_test}
\end{equation}
Here, $h^j, h^k \in \{h^{1}, ..., h^{d_{tr}}\}, j\neq k$, $x^{te}_{i}$ is the target domain sample and $s$ is the softmax function. 
A larger $\mathcal{DS}$ generally means a more dramatic distribution shift. By minimizing $\mathcal{DS}$ we hope to narrow the domain gap and achieve better generalization performance. In the following, we will describe how to leverage $\mathcal{DS}$ for better feature alignment at training and testing stages. 

\subsection{Adversarial Training} 
Given source data from $d_{tr}$ domains, we learn $d_{tr}$ classifiers $\{h^{1}, ..., h^{d_{tr}}\}$ on a shared feature extractor $g$. Each domain-specific classifier $h^{d}$ is first empowered to be discriminative to the corresponding domain $d$ by the simple Empirical Risk Minimization (ERM):
\begin{equation}
    \mathcal{ERM} = \frac{1}{n_{d}}\sum_{i=1}^{n_{d}}\ell(h^{d}(g(x_{i}^{d})), y_{i}^{d}),
\end{equation}
where $\ell$ is the cross-entropy loss. 

With the above notions, a straightforward idea to adapt target domains would be directly applying the trained source classifiers to minimize $\mathcal{DS}$ at test time. 
However, it is generally hard to achieve desired generalization performance, especially when large distribution differences exist among source domains. 
That is because the features of each source domain show large variations in the feature space. 
With unaligned feature distribution, the generalization capability of a specific classifier $h^{d}$ to unseen domain $\{1,...,d_{tr}\}\backslash\{d\}$ is not tightly guaranteed. 
Besides, optimizing classifiers by directly minimizing $\mathcal{DS}$ at training time may lead classifiers to lose domain-specific characteristics and become the same, making minimizing $\mathcal{DS}$ at test time less effective. 

Therefore, we propose an adversarial training strategy which encourages the feature extractor to generate domain-invariant features, and meanwhile, learns distinct domain-specific classifiers. 
Basically, a specific source domain $d$ can be considered as a target domain for classifiers of other source domains $\{h^{1},...,h^{d_{tr}}\}\backslash\{h^{d}\}$. 
Following the terminology of meta-learning~\cite{vanschoren2019meta,balaji2018metareg}, domain $d$ is called the meta-source domain for $h^{d}$, and the meta-target domain for the remaining classifiers. 
Likewise, we define $h^{d}$ as the meta-source classifier for domain $d$ and the meta-target classifier for the other domains (as shown in Fig. \ref{fig:main}(c)). 

Then for a meta-source sample, we minimize ERM loss on both the feature extractor and the meta-source classifier. 
$\mathcal{DS}$ is calculated between any two meta-target classifiers. 
The feature extractor is updated to minimize $\mathcal{DS}$, resulting in the generated features close to the shared feature region, while meta-target classifiers are optimized to maximize $\mathcal{DS}$, avoiding meta-target classifiers become too similar. 
Ideally, after the model converges, we will have the generated features located in the feature regions shared by all involved source domains, and distinct domain-specific classifiers each adapted well to meta-target domains. 
Compared with the naive ERM training which only learns a common classifier on all source domains, multiple domain-specific classifiers will learn various decision boundaries. They together describe a more concrete and invariant distribution for each label class. 

The adversarial training of the feature extractor and classifiers is performed using the gradient reversal layer (GRL)~\cite{ganin2015unsupervised} which reverses the gradient direction to optimize classifiers. As such, we could train all objectives end-to-end. $\mathcal{DS}$ at training time will be rephrased as: 
\begin{equation}
    \mathcal{DS}_{tr} = \sum_{\substack{d,j,k \in \{1,...,d_{tr}\}  \\d\neq j\neq k}}\frac{1}{n_{d}}\sum_{i=1}^{n_{d}}||s(h^{j}(g(x^{d}_{i})))-s(h^{k}(g(x^{d}_{i})))||_1.
    \label{ds_train}
\end{equation}
Note that we exclude the meta-source classifier when computing $\mathcal{DS}$, as it tends to produce overconfident predictions on meta-source data. This would bias the direction to the invariant feature space and make the training hard to optimize. We also add an entropy loss ($\mathcal{ET}$) for regularization which is calculated as: 
\begin{equation}
    \mathcal{ET} = -\frac{1}{C}\sum_{c=1}^{C}log(\frac{1}{n_{d}}\sum_{i=1}^{n_{d}}s(h^{d}(g(x^{d}_{i})))),
    \label{et}
\end{equation}
where $C$ is the number of classes. 

The overall losses are as follows:
\begin{equation}
     \mathop{\arg\min}_{\Phi_{g,\{h^{1},...,h^{d_{tr}}\}}}\sum_{d=1}^{d_{tr}}(\mathcal{ERM}+\gamma \mathcal{ET}) + \lambda \mathop{\arg\min_{\Phi_{g}}\max_{\Phi_{\{h^{1},...,h^{d_{tr}}\}}}}\mathcal{DS}_{tr},
\end{equation}
where $\gamma$ and $\lambda$ are the weights for the entropy and the prediction disagreement-based regularization, $\Phi_{g,\{h^{1},...,h^{d_{tr}}\}}$ are parameters of the feature extractor and classifiers. $\gamma$ is set to 0.01 for all experiments. See Algorithm \ref{alg:crat} for details. 

\begin{algorithm}[htb]
\renewcommand{\algorithmicrequire}{\textbf{Input:}}
\renewcommand{\algorithmicensure}{\textbf{Output:}}
\renewcommand{\algorithmiccomment}{ \ \ \ // }
\caption{AdaODM Training Algorithm}
\begin{algorithmic}[1]
\REQUIRE num of source domains $d_{tr}$, training steps $N$, batch size $B$, gradient ratio $\eta$, weight for the entropy regularization $\gamma$, weight for the disagreement score $\lambda$, the feature extractor $g$, domain-specific classifiers $\{h^{d}\}_{d=1}^{d_{tr}}$, training sets $D^{d}=\{(x^{d}, y^{d})\}_{d=1}^{d_{tr}}, d\in \{1,2,...,d_{tr}\}$ 
\FOR{$step = 1, 2, ..., N$}
    \STATE Draw random data batch $\{(x_{i}^{d}, y_{i}^{d})\}_{i=1}^{B}$ from $D^{d}$
    \STATE Extract features $\{s^{d}_{i}\}_{i=1}^{B}$ by $s_{i}^{d}=g(x_{i}^{d})$ 
    \STATE $\mathcal{L}\_ERM = \mathcal{L}\_DS = \mathcal{L}\_ET = 0$
    \FOR{$d = 1, 2, ..., d_{tr}$}
        \STATE $\mathcal{L}\_ERM  \mathrel{+}=  cross\_entropy((h^{d}(s_{i}^{d}), y_{i}^{d}))$
        \STATE $\mathcal{L}\_Entropy \mathrel{+}= \mathcal{ET}$
        \STATE Reverse gradient directions for $\{h^{1}, h^{2}, ..., h^{d_{tr}}\}\backslash\{h^{d}\}: h.grad=-\eta h.grad$
        \FOR{$(j = 1, 2, ..., d_{tr}) \& j \neq d$}
            \FOR{$(k = 1, 2, ..., d_{tr}) \& (k \neq d) \& (k \neq j)$}
                \STATE $\mathcal{L}\_DS \mathrel{+}= ||softmax(h^{j}(s_{i}^{d})) - softmax(h^{k}(s_{i}^{d}))||_1$
            \ENDFOR
        \ENDFOR
    \ENDFOR
    \STATE $\mathcal{L} = \mathcal{L}\_ERM + \gamma \mathcal{L}\_ET + \lambda \mathcal{L}\_DS$
    \STATE $\Phi_{g}, \Phi_{\{h^{1},h^{2},...,h^{d_{tr}}\}}=SGD(\mathcal{L}, \Phi_{g}, \Phi_{\{h^{1},h^{2},...,h^{d_{tr}}\}})$
\ENDFOR
\end{algorithmic}
\label{alg:crat}
\end{algorithm}

\begin{algorithm}[htb]
\renewcommand{\algorithmicrequire}{\textbf{Input:}}
\renewcommand{\algorithmicensure}{\textbf{Output:}}
\renewcommand{\algorithmiccomment}{ \ \ \ // }
\caption{AdaODM Testing Algorithm}
\begin{algorithmic}[1]
\REQUIRE num of source domains $d_{tr}$, training steps $N$, batch size $B$, trained feature extractor $g$, trained domain-specific classifiers $\{h^{d}\}_{d=1}^{d_{tr}}$, testing sets $D^{d_{te}}=\{x^{d_{te}}\}$, class $c\in \{1,2,...,C\}$ 
\STATE Freeze $\{h^{d}\}_{d=1}^{d_{tr}}$
\STATE Receive data batch $\{x_{i}^{d_{te}}\}_{i=1}^{B}$ from $D^{d_{te}}$
\FOR{$step = 1,2,...,N$}
\STATE Extract features $\{s^{d_{te}}_{i}\}_{i=1}^{B}$ by $s_{i}^{d_{te}}=g(x_{i}^{d_{te}})$
\STATE $\mathcal{L}\_DS = 0$
\FOR{$j = 1, 2, ..., d_{tr}$}
    \FOR{$(k = 1, 2, ..., d_{tr}) \& (k \neq j)$}
        \STATE $\mathcal{L}\_DS \mathrel{+}= ||softmax(h^{j}(s_{i}^{d_{te}})) - softmax(h^{k}(s_{i}^{d_{te}}))||_1$
    \ENDFOR
\ENDFOR
\STATE $\Phi_{g}=SGD(\mathcal{L}\_DS, \Phi_{g})$
\ENDFOR
\STATE Prediction $\hat{y}^{d_{te}}_{i}=\mathop{\arg\max}_{c}(\sum_{d=1}^{d_{tr}}(h^{d}(g(x_{i}^{d_{te}}))))$
\end{algorithmic}
\label{alg:tcr}
\end{algorithm}

\subsection{Online Disagreement Minimization} 

Note that the adversarial training above has enabled a domain-invariant feature extractor and multiple different classifiers. 
This could improve the generalization capacity to an extent. 
However, when faced with query samples that are far from source domains, domain shift still exists. 
Fortunately, the classifiers could effectively capture domain shifts with prediction disagreement. 
Based on this, we propose to adjust the feature extractor with $\mathcal{DS}$ during testing, which allows the model to be optimized for target domains. 
Specifically, we fix domain-specific classifiers and finetune the feature extractor with Eq. \eqref{eq:TTA}. 
\begin{equation} \label{eq:TTA}
    \mathop{\arg\min}_{\Phi}\mathcal{DS},    
\end{equation}
where $\Phi$ represents the parameters of the feature extractor to be updated. Following~\cite{wang2020tent}, we update the scale and shifting parameters of batch-norm layers only. This avoids forgetting knowledge learned at the training stage. As shown in the right of Fig. \ref{fig:main}(d), by minimizing prediction disagreement, we move features of target samples to the domain-invariant feature region, which largely alleviates domain shift in the latent space. See Algorithm \ref{alg:tcr} for the testing stage of AdaODM. Note that AdaODM does not require knowing all test data ahead. Test samples arrive batch by batch and are processed only once. Inference and adaptation are performed simultaneously.

\section{Experiments}
\subsection{Experimental Settings}
\subsubsection{Datasets}
We extensively evaluate AdaODM on four standard domain generalization benchmarks, namely VLCS~\cite{fang2013unbiased}, PACS~\cite{li2017deeper}, OfficeHome~\cite{venkateswara2017deep}, and TerraIncognita~\cite{beery2018recognition} using the protocol of DomainBed~\cite{gulrajani2020search}$\footnote{https://github.com/facebookresearch/DomainBed}$. VLCS comprises four photographic domains $d\in$ \{Caltech101, LabelMe, SUN09, VOC2007$\}$ with 10,729 examples of 5 classes. PACS comprises four domains $d\in$ \{art, cartoons, photos, sketches\} with 9,991 examples of 7 classes. OfficeHome contains four domains $d\in$ \{art, clipart, product, real\} with 15,588 examples of 65 classes. TerraIncognita contains photographs of wild animals taken by camera traps at different locations. Following~\cite{gulrajani2020search}, we use four domains $d\in$ \{L100, L38, L43, L46\} with 24,788 examples of 10 classes. 

\subsubsection{Baselines}
Since AdaODM follows the adaptive domain generalization setting, we compare with both domain generalization and test-time adaptation algorithms. Domain generalization algorithms involve Empirical Risk Minimization (ERM)~\cite{vapnik1999overview}, Invariant Risk Minimization (IRM)~\cite{arjovsky2019invariant}, Group Distributionally Robust Optimization (GroupDRO)~\cite{sagawa2019distributionally}, Inter-domain Mixup (Mixup)~\cite{xu2020adversarial,yan2020improve,wang2020heterogeneous}, Meta-Learning for Domain Generalization (MLDG)~\cite{li2018learning}, Deep CORrelation ALignment (CORAL)~\cite{sun2016deep}, Maximum Mean Discrepancy (MMD)~\cite{li2018domain}, Domain Adversarial Neural Networks (DANN)~\cite{ganin2016domain}, Class-conditional DANN (CDANN)~\cite{li2018deep}, Marginal Transfer Learning (MTL)~\cite{blanchard2011generalizing}, Style-Agnostic Networks (SagNet)~\cite{nam2019reducing}, Adaptive Risk Minimization (ARM)~\cite{zhang2021adaptive}, Variance Risk Extrapolation (VREx)~\cite{krueger2021out}, Representation Self Challenging (RSC)~\cite{huang2020self}, Mixstyle~\cite{zhou2020domain}, mDSDI~\cite{bui2021exploiting}, and Adaptive Domain Generalization (DA-ERM and DA-CORAL)~\cite{dubey2021adaptive}. Note that MixStyle is reproduced number and the others are directly reported from the original paper and DomainBed~\cite{gulrajani2020search}.

We also compare AdaODM with existing test-time adaptation approaches including T3A~\cite{iwasawa2021test}, Tent~\cite{wang2020tent}, SHOT~\cite{liang2020we}, Pseudo Labeling (PL)~\cite{lee2013pseudo}. Note that DomainBed~\cite{gulrajani2020search} freezes all batch-norm layers before fine-tuning on source domains, batch-norm statistics keep unchanged throughout the training. Thus, T-BN~\cite{schneider2020removing} which replaces the batch-norm activation statistics estimated on the training sets with those on the test set, is not suitable for comparison. For the same reason, we compare with a modified version of Tent (Tent-) which only optimizes the batch-norm affine parameters by entropy minimization on predictions of target data.

\subsubsection{Implementation Details} 
AdaODM is model-agnostic and can be applied to any domain-invariant-based domain generalization algorithms. Here we mainly consider ERM and CORAL. Following DomainBed framework~\cite{gulrajani2020search}, for all algorithms on all datasets, we choose ImageNet pre-trained ResNet-50 backbone as the feature extractor and utilize a fully-connected layer for each domain-specific classifier. All configurations of a dataset are considered where one domain is used for test and the others serve as training domains. We split each domain of a dataset into 80\% and 20\% splits. The larger parts are utilized for training and final evaluation, and the smaller ones are used for validation. To achieve a fair comparison, for each algorithm and test domain, we run 20 trials over randomly searched hyperparameters sampled from distributions predefined in~\cite{gulrajani2020search}. Learning rate is selected from $10^{Uniform(-5, -3.5)}$, batch size is selected from $2^{Uniform(3, 5.5)}$, dropout rate is selected from [0, 0.1, 0.5], and weight decay is selected from $10^{Uniform(-6, -2)}$. For introduced $\lambda$ and $\eta$ in our proposed adversarial training procedure, we use $\lambda \sim \mathcal{U}(0,1)$ and $\eta \sim \mathcal{U}(0,2)$, respectively. The above process is repeated for three rounds with different hyperparameters, weight initializations, and dataset splits. The three rounds use seeds $\{0,1,2\}$. For example, there will be totally 240 experiments ($4 \ domains \ \times \ 20 \ trials \ \times \ 3 \ rounds$) for implementing ERM-based AdaODM on OfficeHome. Every result is reported as a mean and standard error over three repetitions. The standard training-domain validation is applied to select the best set of hyperparameters, i.e. smaller splits of each training domain constitute an overall validation set to choose a model. 

To implement TTA baselines, we train source models with ERM and CORAL following DomainBed, respectively. During inference, we fix batch size to 64 for all experiments of both AdaODM and other TTA approaches. The other hyperparameters are set to the same as that during training. For AdaODM, Tent-, SHOT, and PL, two test-time hyperparameters are involved: (1) learning rate, we choose \{0.1, 1.0, 10.0\} times of the learning rate used for training the source model, and (2) steps each test sample is iterated, we choose \{1, 3, 5, 7\}. T3A has one hyperparameter to decide how many supports to store. We test \{1, 5, 20, 50, 100, ALL\} samples. Results are selected similarly as training.

\begin{table*}
    \caption{Domain generalization accuracy for all datasets and algorithms. Values in {\textcolor{myred}{\textbf{red}}} and {\textcolor{myblue}{\textbf{blue}}} indicates optimal and suboptimal performances in TTA or Adaptive DG.}
    \centering
    \adjustbox{max width=\textwidth}{%
    \begin{tabular}{llcccccccc}
    \toprule
    \textbf{Type} & \textbf{Algorithm} & \textbf{VLCS} & \textbf{PACS} & \textbf{OfficeHome} & \textbf{TerraInc} & \textbf{Avg} \\
    \midrule
    \multirow{14}{*}{DG} & ERM~\cite{vapnik1999overview} & 77.5 $\pm$ 0.4 & 85.5 $\pm$ 0.2 & 66.5 $\pm$ 0.3 & 46.1 $\pm$ 1.8 & 69.0 \\
     & IRM~\cite{arjovsky2019invariant} & 78.5 $\pm$ 0.5 & 83.5 $\pm$ 0.8 & 64.3 $\pm$ 2.2 & 47.6 $\pm$ 0.8 & 68.5 \\
     & GroupDRO~\cite{sagawa2019distributionally} & 76.7 $\pm$ 0.6 & 84.4 $\pm$ 0.8 & 66.0 $\pm$ 0.7 & 43.2 $\pm$ 1.1 & 67.6 \\
     & Mixup~\cite{xu2020adversarial,yan2020improve,wang2020heterogeneous} & 77.4 $\pm$ 0.6 & 84.6 $\pm$ 0.6 & 68.1 $\pm$ 0.3 & 47.9 $\pm$ 0.8 & 69.5 \\
     & MLDG~\cite{li2018learning} & 77.2 $\pm$ 0.4 & 84.9 $\pm$ 1.0 & 66.8 $\pm$ 0.6 & 47.7 $\pm$ 0.9 & 69.2 \\
     & CORAL~\cite{sun2016deep} & 78.8 $\pm$ 0.6 & 86.2 $\pm$ 0.3 & 68.7 $\pm$ 0.3 & 47.6 $\pm$ 1.0 & 70.3 \\
     & MMD~\cite{li2018domain} & 77.5 $\pm$ 0.9 & 84.6 $\pm$ 0.5 & 66.3 $\pm$ 0.1 & 42.2 $\pm$ 1.6 & 67.7 \\
     & DANN~\cite{ganin2016domain} & 78.6 $\pm$ 0.4 & 83.6 $\pm$ 0.4 & 65.9 $\pm$ 0.6 & 46.7 $\pm$ 0.5 & 68.7 \\
     & CDANN~\cite{li2018deep} & 77.5 $\pm$ 0.1 & 82.6 $\pm$ 0.9 & 65.8 $\pm$ 1.3 & 45.8 $\pm$ 1.6 & 67.9 \\
     & MTL~\cite{blanchard2011generalizing} & 77.2 $\pm$ 0.4 & 84.6 $\pm$ 0.5 & 66.4 $\pm$ 0.5 & 45.6 $\pm$ 1.2 & 68.5 \\
     & SagNet~\cite{nam2019reducing} & 77.8 $\pm$ 0.5 & 86.3 $\pm$ 0.2 & 68.1 $\pm$ 0.1 & 48.6 $\pm$ 1.0 & 70.2 \\
     & ARM~\cite{zhang2021adaptive} & 77.6 $\pm$ 0.3 & 85.1 $\pm$ 0.4 & 64.8 $\pm$ 0.3 & 45.5 $\pm$ 0.3 & 68.3 \\
     & VREx~\cite{krueger2021out} & 78.3 $\pm$ 0.2 & 84.9 $\pm$ 0.6 & 66.4 $\pm$ 0.6 & 46.4 $\pm$ 0.6 & 69.0 \\
     & RSC~\cite{huang2020self} & 77.1 $\pm$ 0.5 & 85.2 $\pm$ 0.9 & 65.5 $\pm$ 0.9 & 46.6 $\pm$ 1.0 & 68.6 \\
     & MixStyle~\cite{zhou2020domain} & 77.8 $\pm$ 0.3 & 85.3 $\pm$ 0.4 & 61.3 $\pm$ 0.7 & 44.5 $\pm$ 1.1 & 67.2 \\
     & mDSDI~\cite{bui2021exploiting} & 79.0 $\pm$ 0.3 & 86.2 $\pm$ 0.2 & 69.2 $\pm$ 0.4 & 48.1 $\pm$ 1.4 & 70.6 \\
     & DA-ERM~\cite{dubey2021adaptive} & 78.0 $\pm$ 0.2 & 84.1 $\pm$ 0.5 & 67.9 $\pm$ 0.4 & 47.3 $\pm$ 0.5 & 69.3 \\
     & DA-CORAL~\cite{dubey2021adaptive} & 78.5 $\pm$ 0.4 & 84.5 $\pm$ 0.7 & 68.9 $\pm$ 0.4 & 48.1 $\pm$ 0.3 & 70.0 \\
    \midrule
    \multirow{5}{*}{TTA} & ERM (Vanilla Training) & 78.1 $\pm$ 0.6 & 84.7 $\pm$ 0.9 & 66.8 $\pm$ 0.4 & 46.9 $\pm$ 1.1& 69.1 \\
     & +T3A~\cite{iwasawa2021test} & {\textcolor{myred}{\textbf{79.9 $\pm$ 1.6}}} & 85.5 $\pm$ 0.9 & 68.5 $\pm$ 0.5 &  47.2 $\pm$ 0.7 & 70.3 \\ 
     & +Tent-~\cite{wang2020tent} & 77.1 $\pm$ 1.5 & 85.4 $\pm$ 0.9 & 65.6 $\pm$ 0.3 & 44.4 $\pm$ 1.1 & 68.1 \\
     & +SHOT~\cite{liang2020we} & 72.7 $\pm$ 0.9 & 85.5 $\pm$ 0.9 & 66.9 $\pm$ 0.2 & 37.0 $\pm$ 1.0 & 65.5 \\
     & +PL~\cite{lee2013pseudo} & 73.5 $\pm$ 3.6& 84.7 $\pm$ 0.5 & 61.5 $\pm$ 3.5 & 40.6 $\pm$ 0.6 & 65.1 \\
    \midrule
    \multirow{5}{*}{TTA} & CORAL (Vanilla Training) & 77.8 $\pm$ 0.7 & 85.2 $\pm$ 0.3 & 68.9 $\pm$ 0.2 & 46.3 $\pm$ 0.5 & 69.6 \\
     & +T3A~\cite{iwasawa2021test} & 78.4 $\pm$ 0.9 & 86.2 $\pm$ 0.8 & 69.4 $\pm$ 0.5 & 46.8 $\pm$ 2.0 & 70.2 \\
     & +Tent-~\cite{wang2020tent} & 76.1 $\pm$1.0 & 85.4 $\pm$ 0.8 & 68.6 $\pm$ 1.4 & {\textcolor{myblue}{\textbf{47.5 $\pm$ 2.3}}} & 69.4 \\
     & +SHOT~\cite{liang2020we} & 74.4 $\pm$ 3.2& 86.3 $\pm$ 0.8 & {\textcolor{myblue}{\textbf{69.6 $\pm$ 0.5}}} & 39.0 $\pm$ 1.1 & 67.3 \\
     & +PL~\cite{lee2013pseudo} & 72.6 $\pm$ 3.4 & 84.0 $\pm$ 0.4 & 66.1 $\pm$ 3.7& 47.0 $\pm$ 1.4 & 67.4 \\
    \midrule
    \multirow{2}{*}{\makecell[l]{Adaptive DG \\ (Ours)}} & ERM (AdaODM Training) & 77.7 $\pm$ 0.5 & 84.1 $\pm$ 0.2 & 67.2 $\pm$ 0.3 & 46.8 $\pm$ 1.1 & 69.0 \\
     & +AdaODM Testing & {\textcolor{myblue}{\textbf{79.7 $\pm$ 0.8}}} & {\textcolor{myred}{\textbf{87.8 $\pm$ 0.5}}} & 68.8 $\pm$ 0.2 & {\textcolor{myred}{\textbf{47.7 $\pm$ 1.2}}} & {\textcolor{myred}{\textbf{71.0}}} \\  
    \midrule
    \multirow{2}{*}{\makecell[l]{Adaptive DG \\ (Ours)}} & CORAL (AdaODM Training) & 77.1 $\pm$ 0.5 & 85.4 $\pm$ 0.6 & 68.4 $\pm$ 0.0 & 46.2 $\pm$ 3.7 & 69.3 \\
     & +AdaODM Testing & 79.1 $\pm$ 1.1 & {\textcolor{myblue}{\textbf{86.6 $\pm$ 1.4}}} & {\textcolor{myred}{\textbf{69.9 $\pm$ 0.2}}} & 47.5 $\pm$ 1.8 & {\textcolor{myblue}{\textbf{70.8}}}\\    
    \bottomrule
    \end{tabular}}
    \label{tab1}

\end{table*}

\subsection{Comparison with Existing DG Methods}
We first compare with existing domain generalization methods. Test accuracies are summarized in Table \ref{tab1}. Table \ref{tab1} is split into five blocks. The first block reports the performance of the existing domain generalization algorithms (the first block from \textsl{ERM} to \textsl{DA-CORAL}). The fourth and the fifth blocks are results of AdaODM, built on ERM and CORAL, respectively. Within each of these two blocks, performance of both base models (directly inference with source models) and adapted models (via test-time online disagreement minimization) are posted. Notably, our model involves multiple classifiers, we aggregate multiple domain-specific classifiers to obtain the final predictions. 

Firstly, by comparing the performances of base models trained with Prediction Disagreement-based Adversarial Training (PD-AT) and those trained in a vanilla manner, both ERM and CORAL show no significant differences. It is reasonable since PD-AT applied in our training process builds a feature space invariant only to source domains. The feature extractor cannot guarantee that target features are mapped into the shared feature clusters. Therefore, inference directly using base models gets limited benefits. However, significant performance gains are achieved after conducting test-time adaptive DG. AdaODM at test time stably improves the performances of ERM by 2.0\%, 3.7\%, 1.6\%, and 0.9\% on VLCS, PACS, OfficeHome, and TerraIncognita, respectively. Similarly, we consistently improve the performance of CORAL on all four datasets, by 2.0\%, 1.2\%, 1.5\%, and 1.3\%, respectively. Notably, although the average performance of our reproduced CORAL (69.3\%) is a little lower than the reported score (70.3\%), we still achieve a higher performance of 70.8\% after test-time adaptive DG. In particular, the proposed method achieves 87.8\% for PACS and 69.9\% for OfficeHome, which both surpass the best-reported scores. Overall, we achieve state-of-the-art performance of 71.0\% with ERM in terms of the average test accuracy. This indicates that during inference, by encouraging prediction consistency, the feature extractor effectively aligns target features to the domain-invariant feature space. 

\begin{figure*}
	\centering
	\subfloat[Target domain: Clipart]{
	\begin{minipage}[b]{0.5\textwidth}
        \captionsetup[subfloat]{labelformat=empty}
		\subfloat[w/o AT]{
		  \includegraphics[width=0.46\textwidth]{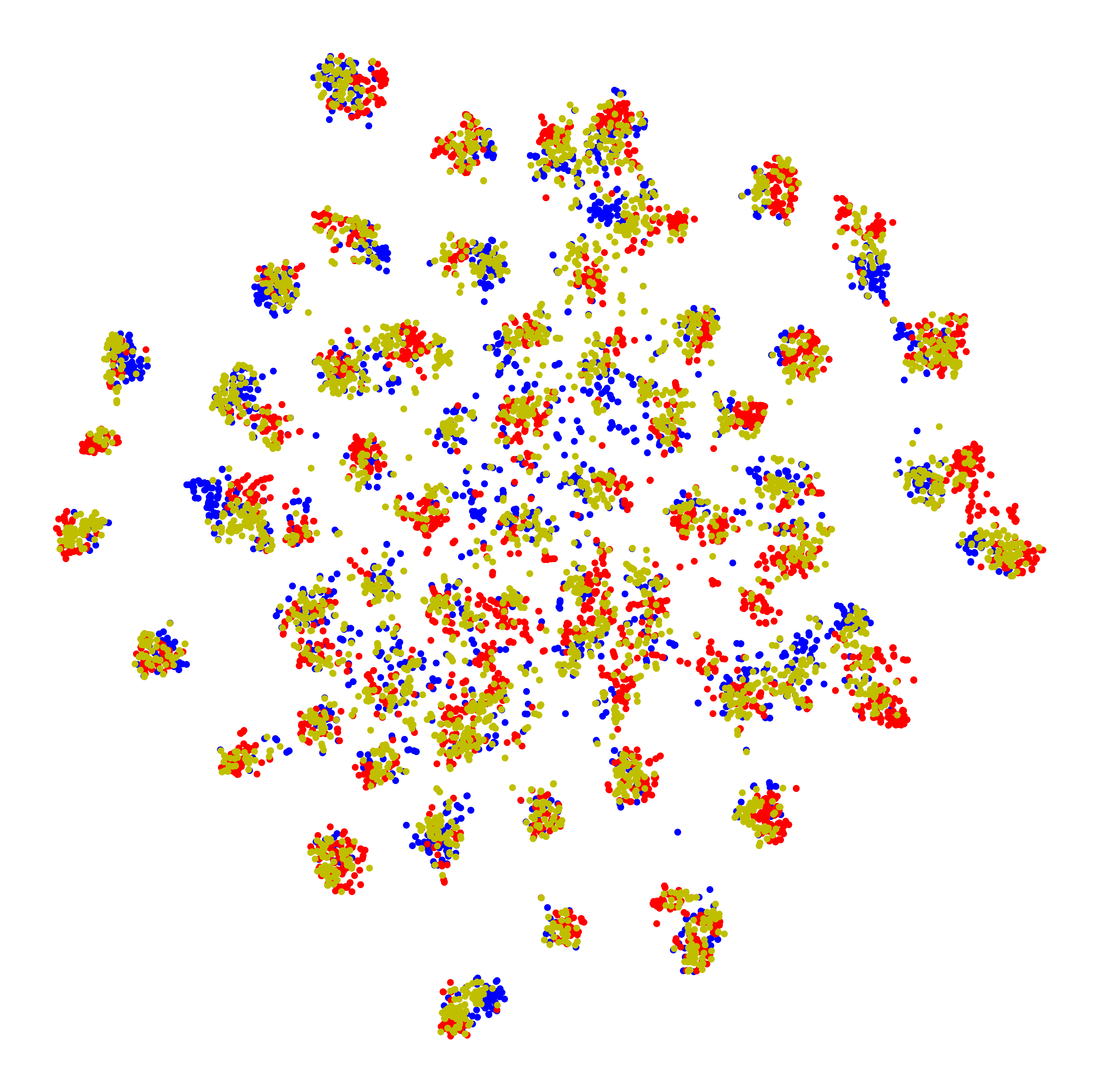}
        }
        \setcounter{subfigure}{0}
	   \subfloat[w/ AT]{
	       \includegraphics[width=0.46\textwidth]{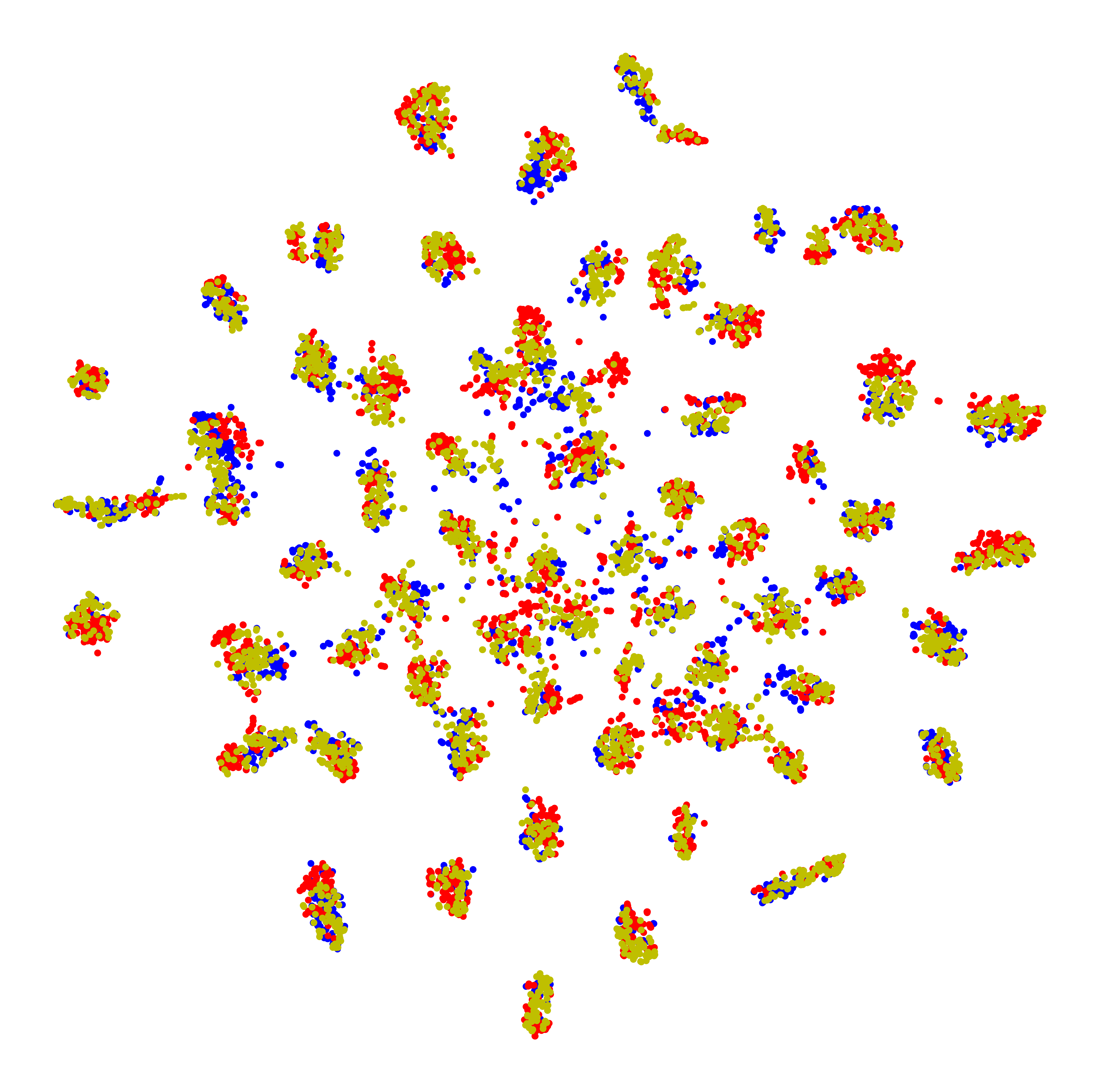}
        }
        
	\end{minipage}
	}	
	\subfloat[Target domain: Art]{
	\begin{minipage}[b]{0.5\textwidth}
	    \setcounter{subfigure}{0}
	    \captionsetup[subfloat]{labelformat=empty}
		\subfloat[w/o AT]{ 
		  \includegraphics[width=0.46\textwidth]{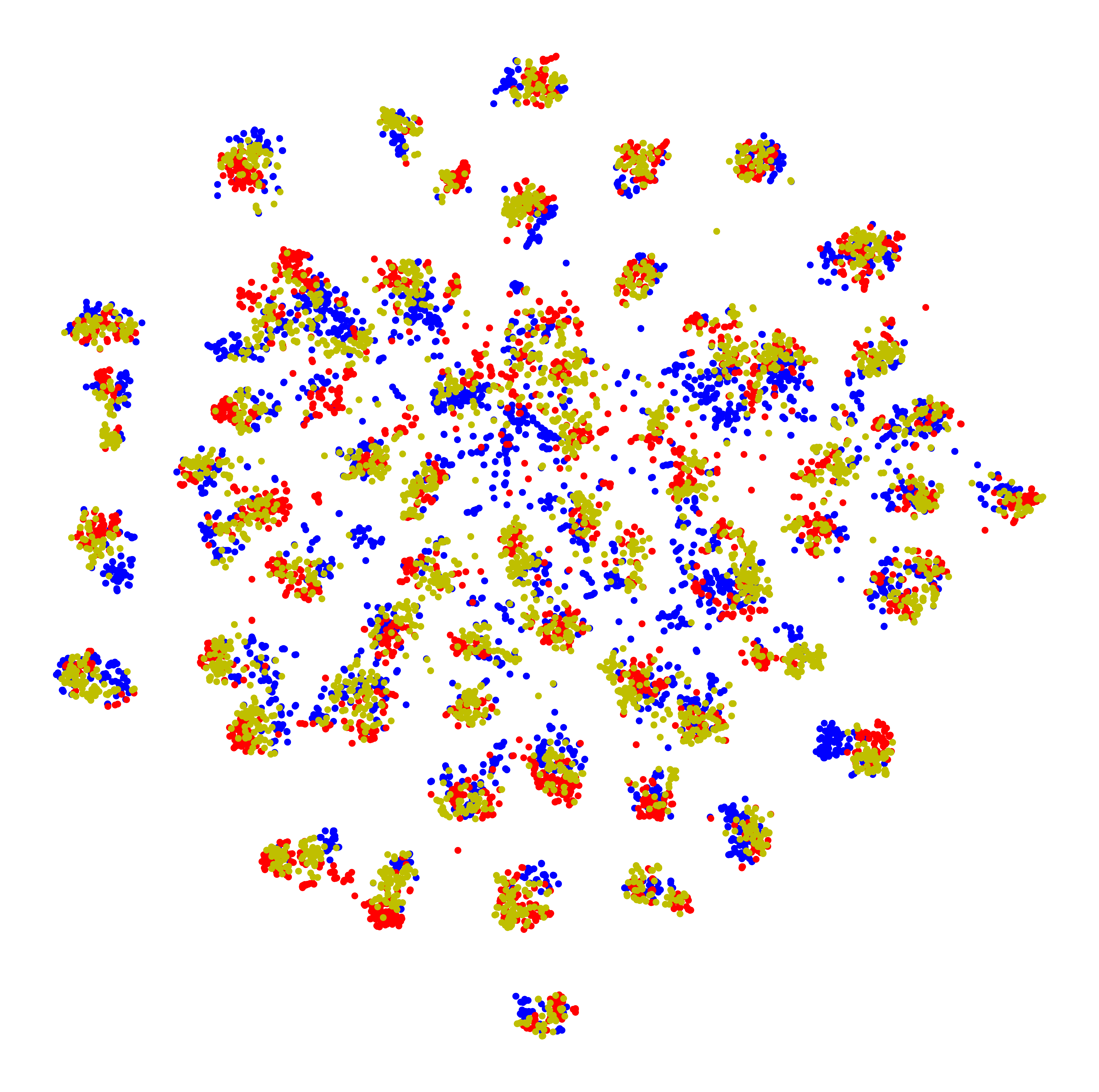}
        }
	    \subfloat[w/ AT]{
	       \includegraphics[width=0.46\textwidth]{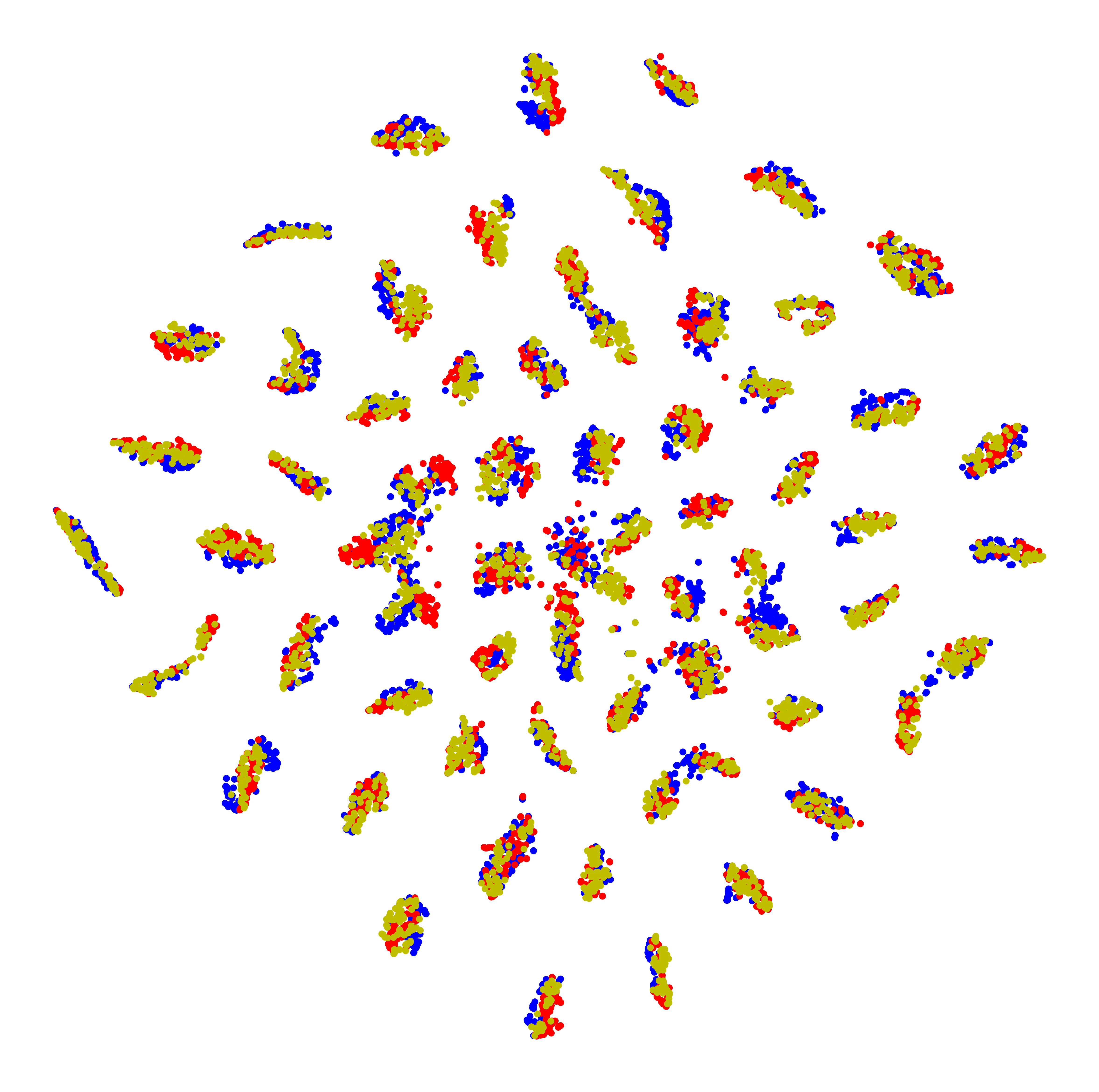}
        }
	\end{minipage}
	}
    \caption{Two examples to illustrate the effectiveness of the prediction disagreement-based adversarial training. The left and the right pairs are tested on ``clipart'' and ``art'' domains of the OfficeHome dataset, respectively, and trained on the other three domains. Within each pair, the two figures visualize generated source domain features trained without and with the \textbf{Adversarial Training (AT)}. Three colors represent three source domains. We can see that AT effectively encourages domain-invariant feature clusters. }
\label{fig:tnse}
\end{figure*}

From another perspective, we take OfficeHome as an example to visualize the effect of the adversarial training in Fig. \ref{fig:tnse}. Specifically, we train vanilla ERM models (without prediction disagreement-based adversarial training) and ERM-based AdaODM models (with prediction disagreement-based adversarial training). Then we visualize the learned features using t-SNE~\cite{van2008visualizing}. The left pair shows an example that trains the source model on ``art'', ``product'', and ``real'' domains, and generalizes to ``clipart'' domain, while the right pair tests on ``art'' domain and trains on the other domains. Within each pair, the two figures illustrate the source features that are generated without and with adversarial training, respectively. It could be easily observed that the adversarial training procedure successfully clusters features from various domains. To evaluate whether the test-time online disagreement minimization benefits the target feature alignment, we apply a CORAL-like metric as an indicator to measure the distance between the learned source and target feature distributions. The metric is calculated as the second-order difference of mean and covariance values between source and target features. For the above two trials, the test-time optimization process of AdaODM improves the test accuracy from 54.1\% to 59.4\%, and from 61.7\% to 64.0\%, respectively. And the metric decreases from 9.07 to 5.30, and from 1.43 to 1.39, respectively. This implies our test-time online disagreement minimization indeed mitigates the distribution gap between source and target domains.

\subsection{Comparison with Existing TTA Methods}
We further compare AdaODM with test-time adaptation methods, including T3A, Tent-, SHOT, and PL. AdaODM is built on source models with multiple domain-specific classifiers, which is different from baseline TTA methods. To make a fair comparison, we train source models separately for AdaODM and existing TTA methods, i.e., we train ERM and CORAL on vanilla architectures with one feature extractor and one classifier for baseline TTA methods, and train ERM and CORAL on architectures with multiple domain-specific classifiers via PD-AT for our test-time adaptive DG. T3A adjusts a pseudo-prototypical classifier with target features to adapt new domains. Tent- optimizes batch-norm transformations with prediction entropy minimization. SHOT freezes the classifier and updates the feature extractor with prediction entropy minimization, diversity regularization, and cross-entropy minimization between prediction and the pseudo label. PL optimizes the whole network with the pseudo label. The pseudo labels in SHOT and PL are both determined by thresholding the prediction probabilities with 0.9. T3A, Tent-, SHOT, and PL are conducted based on the same base models. All base models are selected based on the training-domain validation standard from 20 $\times$ 3 training models for each test environment. 

The second and the third blocks in Table \ref{tab1} summarize performances of baseline TTA methods with ERM and CORAL, respectively. It can be observed that among all baselines, only T3A gets improved performance over all datasets. SHOT performs better than base models on PACS and OfficeHome. Tent- receives better performance on PACS. PL obtains poor performance on most datasets. However, our method consistently outperforms T3A or achieves comparable performances. In particular, our method surpasses T3A by 0.7\% with ERM and 0.5\% with CORAL. 
Entropy minimization used in SHOT and Tent- benefits adaptation on several datasets, but fails on the others. Minimizing prediction entropy prompts more confident supervision, and might be effective in some circumstances. However, the optimization direction cannot be guaranteed to the right class. Compared to existing TTA methods, our method first builds a domain-invariant feature space by imposing consistency regularization. With the same procedure during inference, target features can also be aligned to the shared feature space. Meanwhile, domain-specific classifiers could better measure the distribution difference between the target domain and source domains. Both factors contribute to the reliability of the optimization direction at test time. 

\subsection{Quantitative analysis of adaptive DG results}
\begin{figure*}
	\centering
	\subfloat[CORAL-based AdaODM on OfficeHome]{
		\begin{minipage}[b]{0.30\textwidth}
			\includegraphics[width=0.9\textwidth]{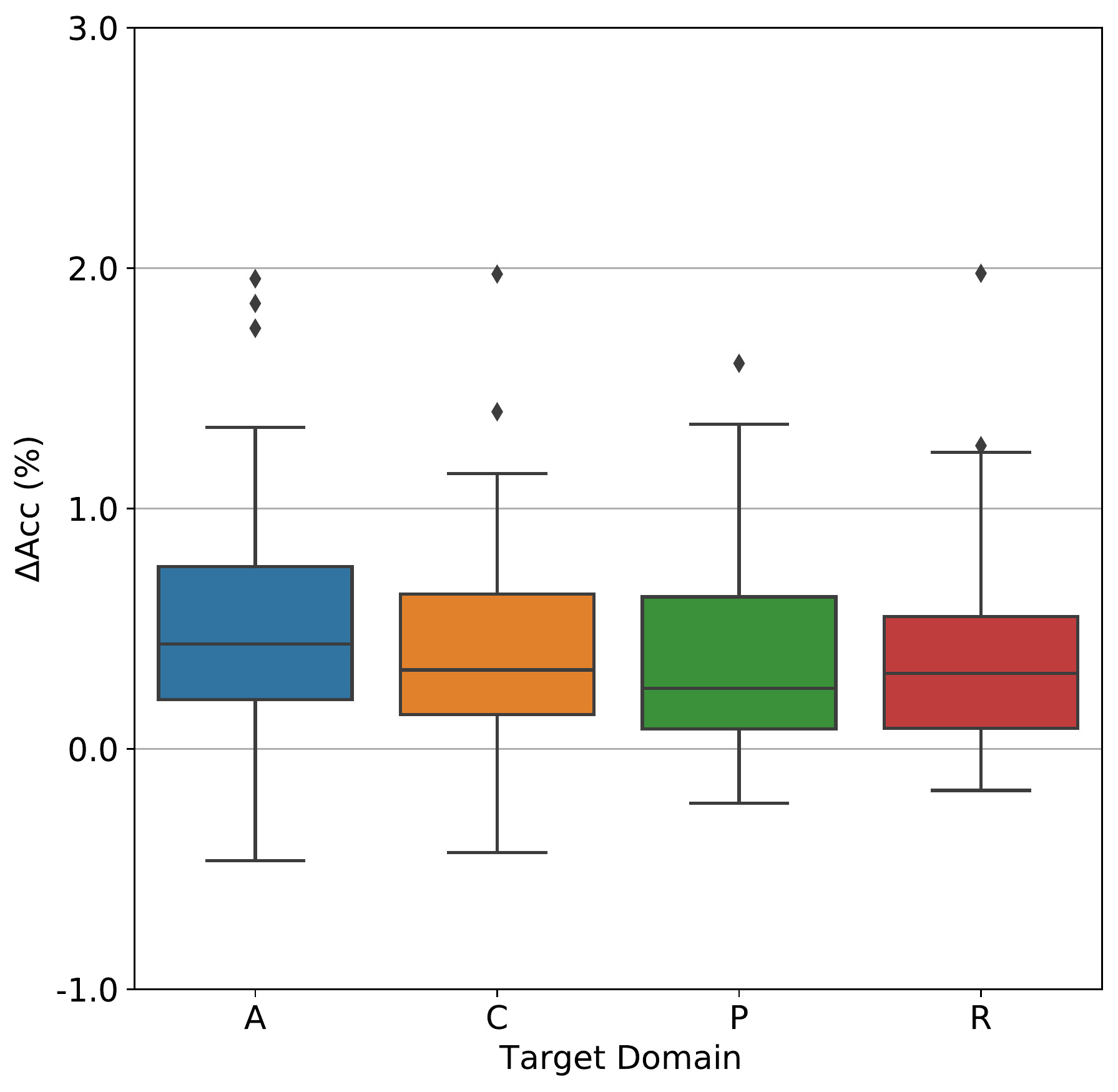}
		\end{minipage}
		\label{fig:box_OH}
	}
	\subfloat[CORAL-based AdaODM on PACS]{
		\begin{minipage}[b]{0.30\textwidth}
  	 	\includegraphics[width=0.9\textwidth]{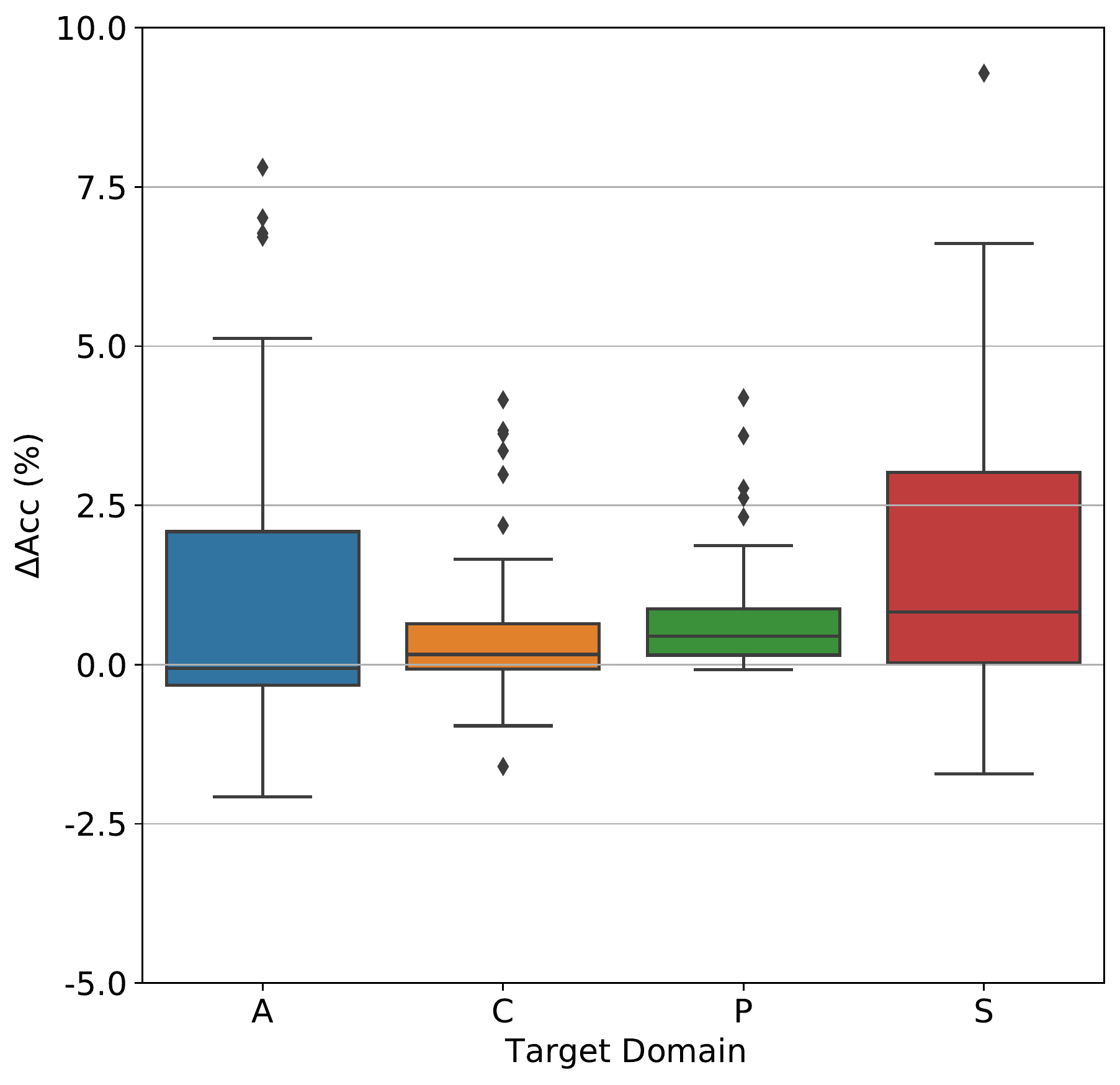}
		\end{minipage}
	\label{fig:box_PACS}
	}
	\subfloat[CORAL-based AdaODM on TerraIncognita]{
		\begin{minipage}[b]{0.30\textwidth}
  	 	\includegraphics[width=0.9\textwidth]{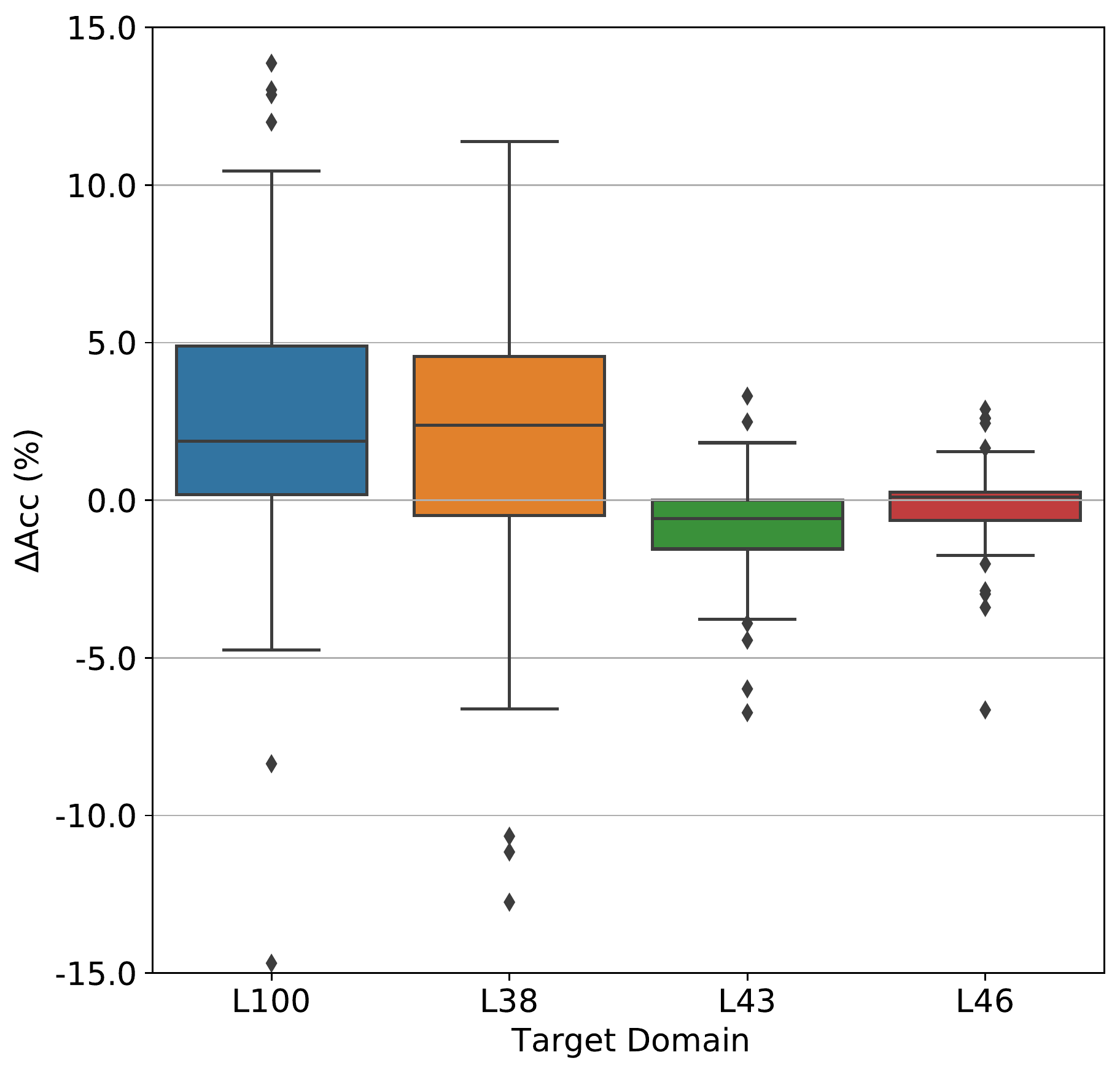}
		\end{minipage}
	\label{fig:box_Terra}
	}
\caption{Distribution of performance gains by CORAL-based AdaODM with different hyperparameters ($20 \ trials \times 3 \ rounds$ for each target domain) on OfficeHome, PACS, and TerraIncognita using box plot.}
\label{fig:box}
\end{figure*}

We evaluate how much the adaptive DG procedure (test-time online disagreement minimization) of AdaODM could boost the target domain performance. Since we have run 60 experiments ($20 \ trials \ \times \ 3 \ rounds$) for each target domain with different hyperparameters and dataset splits, performance gain brought by the test-time online disagreement minimization for each experiment could be calculated. Fig. \ref{fig:box} summarizes performance gains of each domain of OfficeHome, PACS, and TerraIncognita with CORAL-based AdaODM using box plots. We observe that for OfficeHome, the third-quarter quantile of the performance gain is larger than 0 for all target domains, which means AdaODM is robust to the hyperparameters of the base model. On PACS and TerraIncognita, performance gains exhibit more volatile fluctuations. AdaODM could improve the performance of base models by more than 5\% with suitable hyperparameters. AdaODM does not work quite well when generalizing to the ``art'' domain of PACS, ``L43'', and ``L46'' domains of TerraIncognita. 

\subsection{AdaODM with single source domain}
Since domain-specific classifiers in AdaODM are built based on the number of source domains, we are also interested in how AdaODM behaves when there is only a single source domain. Under such a circumstance, we instead construct classifiers with different initialized weights to improve the diversity. Thus, the number of classifiers is independent of the number of source domains. Specifically, we learn 3 different classifiers on a single source domain and implement AdaODM based on ERM. We consider all generalization pairs of each dataset. Every pair is conducted once with default hyperparameters. Table \ref{tab_single} describes the detailed results. We could observe that even with only one source domain, our method still improves the performance at testing stage by 1.4\%, 1.0\%, 1.5\%, and 1.6\% in average for VLCS, PACS, OfficeHome, and TerraIncognita, respectively. 
\begin{table*}
    \caption{Performance of AdaODM for single source domain generalization. \textbf{Bold} numbers indicate performance improvements from base models. Note that experiments are conducted only on the default hyperparameters of ERM-based AdaODM.}
    \centering
    \adjustbox{max width=\textwidth}{%
    \begin{tabular}{l|ccc|ccc|ccc|ccc|c}
    \toprule
    \textbf{VLCS} & \textbf{C$\xrightarrow{}$L} & \textbf{C$\xrightarrow{}$S} & \textbf{C$\xrightarrow{}$V} & \textbf{L$\xrightarrow{}$C} & \textbf{L$\xrightarrow{}$S} & \textbf{L$\xrightarrow{}$V} & \textbf{S$\xrightarrow{}$C} & \textbf{S$\xrightarrow{}$L} & \textbf{S$\xrightarrow{}$V} & \textbf{V$\xrightarrow{}$C} & \textbf{V$\xrightarrow{}$L} & \textbf{V$\xrightarrow{}$S} & \textbf{Avg} \\
    AdaODM Training & 48.0 & 50.0 & 57.8 & 31.4 & 59.2 & 62.9 & 81.0 & 64.3 & 70.4 & 93.2 & 62.7 & 75.1 & 63.0 \\
    +AdaODM Testing & \textbf{48.4} & \textbf{50.6} & 57.5 & \textbf{32.6} & 58.6 & \textbf{63.5} & \textbf{89.3} & \textbf{67.1} & \textbf{72.1} & \textbf{94.3} & 62.2 & \textbf{76.2} & \textbf{64.4} \\
    \midrule
    \textbf{PACS} & \textbf{A$\xrightarrow{}$C} & \textbf{A$\xrightarrow{}$P} & \textbf{A$\xrightarrow{}$S} & \textbf{C$\xrightarrow{}$A} & \textbf{C$\xrightarrow{}$P} & \textbf{C$\xrightarrow{}$S} & \textbf{P$\xrightarrow{}$A} & \textbf{P$\xrightarrow{}$C} & \textbf{P$\xrightarrow{}$S} & \textbf{S$\xrightarrow{}$A} & \textbf{S$\xrightarrow{}$C} & \textbf{S$\xrightarrow{}$P} & \textbf{Avg} \\
    AdaODM Training & 74.6 & 94.9 & 70.4 & 75.0 & 90.6 & 71.4 & 51.2 & 49.1 & 56.6 & 64.3 & 69.4 & 62.6 & 69.2 \\
    +AdaODM Testing & \textbf{76.0} & \textbf{96.8} & \textbf{74.4} & \textbf{76.8} & \textbf{91.5} & \textbf{73.4} & \textbf{51.5} & 48.9 & 56.4 & \textbf{64.9} & 69.4 & 61.8 & \textbf{70.2}\\
    \midrule
    \textbf{OfficeHome} & \textbf{A$\xrightarrow{}$C} & \textbf{A$\xrightarrow{}$P} & \textbf{A$\xrightarrow{}$R} & \textbf{C$\xrightarrow{}$A} & \textbf{C$\xrightarrow{}$P} & \textbf{C$\xrightarrow{}$R} & \textbf{P$\xrightarrow{}$A} & \textbf{P$\xrightarrow{}$C} & \textbf{P$\xrightarrow{}$R} & \textbf{R$\xrightarrow{}$A} & \textbf{R$\xrightarrow{}$C} & \textbf{R$\xrightarrow{}$P} & \textbf{Avg} \\
    AdaODM Training & 44.2 & 57.2 & 68.2 & 44.2 & 54.9 & 56.7 & 40.3 & 38.3 & 64.2 & 57.8 & 46.1 & 70.9 & 53.6 \\
    +AdaODM Testing & \textbf{46.9} & \textbf{58.2} & \textbf{69.3} & \textbf{45.7} & \textbf{55.5} & \textbf{58.2} & \textbf{42.5} & \textbf{39.1} & \textbf{66.1} & \textbf{59.4} & 46.0 & \textbf{74.2} & \textbf{55.1} \\
    \midrule
    \textbf{TerraInc} & \textbf{L100$\xrightarrow{}$L38} & \textbf{L100$\xrightarrow{}$L43} & \textbf{L100$\xrightarrow{}$L46} & \textbf{L38$\xrightarrow{}$L100} & \textbf{L38$\xrightarrow{}$L43} & \textbf{L38$\xrightarrow{}$L46} & \textbf{L43$\xrightarrow{}$L100} & \textbf{L43$\xrightarrow{}$L38} & \textbf{L43$\xrightarrow{}$L46} & \textbf{L46$\xrightarrow{}$L100} & \textbf{L46$\xrightarrow{}$L38} & \textbf{L46$\xrightarrow{}$P} & \textbf{Avg} \\
    AdaODM Training & 16.2 & 29.4 & 38.8 & 34.2 & 21.9 & 21.8 & 41.7 & 44.7 & 39.2 & 40.2 & 45.3 & 54.8 & 35.7 \\
    +AdaODM Testing & \textbf{17.0} & 29.1 & \textbf{39.2} & \textbf{34.6} & \textbf{22.3} & \textbf{24.1} & \textbf{48.4} & \textbf{46.0} & \textbf{39.5} & 38.9 & \textbf{53.8} & \textbf{55.0} & \textbf{37.3}\\
    \bottomrule
    \end{tabular}}
    \label{tab_single}

\end{table*}

\begin{table*}
    \caption{Ablation studies to prove the effectiveness of domain-specific classifiers and adversarial training. \textbf{Bold} numbers indicate best results in each setting.}
    \centering
    \adjustbox{max width=\textwidth}{%
    \begin{tabular}{llcccccc}
    \toprule
    \textbf{Base DG Method} & \textbf{Algorithm} & \textbf{VLCS} & \textbf{PACS} & \textbf{OfficeHome} & \textbf{TerraInc} & \textbf{Avg} \\
    \midrule
    \multirow{3}{*}{ERM} & full & \textbf{79.7 $\pm$ 0.8} & \textbf{87.8 $\pm$ 0.5} & \textbf{68.8 $\pm$ 0.2} & \textbf{47.7 $\pm$ 1.2} & \textbf{71.0} \\
     & w/o domain-specific classifiers & 77.7 $\pm$ 1.0 & 84.7 $\pm$ 1.4 & 66.5 $\pm$ 1.0 & 46.2 $\pm$ 0.6 & 68.8 \\
     & w/o adversarial training & 77.6 $\pm$ 2.0 & 85.9 $\pm$ 0.6 & 67.1 $\pm$ 0.7 & 45.6 $\pm$ 6.4 & 69.1 \\
    \midrule
    \multirow{3}{*}{CORAL} & full & \textbf{79.1 $\pm$ 1.1} & \textbf{86.6 $\pm$ 1.4} & \textbf{69.9 $\pm$ 0.2} & \textbf{47.5 $\pm$ 1.8} & \textbf{70.8} \\
     & w/o domain-specific classifiers  & 78.6 $\pm$ 0.3 & 86.6 $\pm$ 0.5 & 69.4 $\pm$ 0.7 & 47.1 $\pm$ 1.4 & 70.4 \\
     & w/o adversarial training  & 77.7 $\pm$ 0.3 & 86.3 $\pm$ 0.8 & 68.7 $\pm$ 0.2 & 47.3 $\pm$ 3.6 & 70.0 \\
    \bottomrule
    \end{tabular}}
    \label{tab2}
\end{table*}

\begin{table}
    \caption{The effect of the number of source domains.}
    \centering
    \adjustbox{max width=\linewidth }{%
    \begin{tabular}{ccccccc}
    \toprule
    \textbf{Num. of } & & \multirow{2}{*}{\textbf{VLCS}} & \multirow{2}{*}{\textbf{PACS}} & \multirow{2}{*}{\textbf{OfficeHome}} & \multirow{2}{*}{\textbf{TerraInc}} & \multirow{2}{*}{\textbf{Avg}} \\
    \textbf{src. domains} \\
    \midrule
    \multirow{2}{*}{2} & AdaODM Training & 64.5 & 71.3 & 58.3 & 35.5 & 57.4 \\
    & +AdaODM Testing & 64.8 & 71.6 & 58.3 & 36.4 & 57.8 \\
    \multirow{2}{*}{3} & AdaODM Training & 75.0 & 83.5 & 65.6 & 41.9 & 66.5 \\
    & +AdaODM Testing & 75.6 & 84.0 & 66.1 & 42.3 & 67.0 \\
    \bottomrule
    \end{tabular}}
    \label{num_of_domain}
\end{table}

\subsection{The effect of the number of source domains}
We evaluate how the number of source domains affects the generalization performance. Since VLCS, PACS, OfficeHome, and TerraIncognita each contain four domains, we consider two or three source domains used for training and the remaining one domain serves as the target domain. To make a fair comparison, 60\% samples of each source domain are included for training when using three source domains. When the number of source domains is two, the ratio of included images for each source domain is determined accordingly to roughly ensure the total number of training samples is the same. For the case where the number of source domains is two, we report the average performance of sampling any two from the three source domains. When there are two source domains present for training, the number of ``meta-target'' classifiers is only 1 for each source domain. Therefore, we build two differently initialized domain-specific classifiers for each source domain to calculate the prediction disagreement score. 

For each dataset, the final results are obtained by averaging over the four scenarios where each domain serves as the target domain. All experiments are conducted on the default hyperparameters of ERM-based AdaODM. From Table \ref{num_of_domain} we can see that when decreasing the number of source domains from 3 to 2, performance drops heavily: the results of vanilla inference drop from 66.5\% to 57.4\%. However, we still 
observe the performance gain brought by test-time training. We also notice that the gains on each dataset become more fluctuated when decreasing the number of source domains, i.e., results on VLCS, PACS, and OfficeHome are less boosted while TerraIncognita gets higher improvement by test-time optimization.

\subsection{Choice to calculate $\mathcal{DS}$\label{sec:DS}}
Table \ref{tab_DS} compares test accuracies of computing $\mathcal{DS}$ by $L_1$- and $L_2$-norm, and $KL$ Divergence. Experiments are conducted once on each target domain of each dataset with default hyperparameters. ERM-based AdaODM is implemented. Results empirically show that the adaptive DG benefits much more from utilizing $L_1$-norm to calculate $\mathcal{DS}$. This is understandable since $L_1$-norm encourages prediction differences for each class to be more sparse. Thus, models focus more on certain classes. Since classifiers are prone to be less reliable at the early training stage, both $L_2$-norm and $KL$ Divergence may lead meta-target classifiers to agree on wrong predicted distributions. 

\begin{table}
    \caption{Comparison of choices for computing $\mathcal{DS}$. \textbf{Bold} numbers indicate performance improvements from base models. Note that experiments are conducted only on the default hyperparameters of ERM-based AdaODM.}
    \centering
    \adjustbox{max width=\linewidth }{%
    \begin{tabular}{clccccc}
    \toprule
    \textbf{Choice for $\mathcal{DS}$} & & \textbf{VLCS} & \textbf{PACS} & \textbf{OfficeHome} & \textbf{TerraInc} & \textbf{Avg} \\
    \midrule
    \multirow{2}{*}{$L_1$} & AdaODM Training & 75.7 & 84.4 & 65.6 & 41.9 & 66.9 \\
     & +AdaODM Testing & 75.7 & \textbf{85.1} & \textbf{66.9} & \textbf{42.8} & \textbf{67.6} \\
    \midrule
    \multirow{2}{*}{$L_2$} & AdaODM Training & 77.2 & 67.6 & 64.0 & 42.9 & 62.9 \\
     & +AdaODM Testing & 76.9 & \textbf{68.8} & 63.6 & 42.8 & \textbf{63.0} \\
    \midrule
    \multirow{2}{*}{$KL$} & AdaODM Training & 73.2 & 75.1 & 64.0 & 41.3 & 63.4 \\
     & +AdaODM Testing & 72.7 & \textbf{75.5} & 63.9 & 41.0 & 63.3 \\
    \bottomrule
    \end{tabular}}
    \label{tab_DS}
\end{table}

\begin{figure}
	\centering
	\subfloat[VLCS]{
		\begin{minipage}[b]{0.24\textwidth}
			\includegraphics[width=0.9\textwidth]{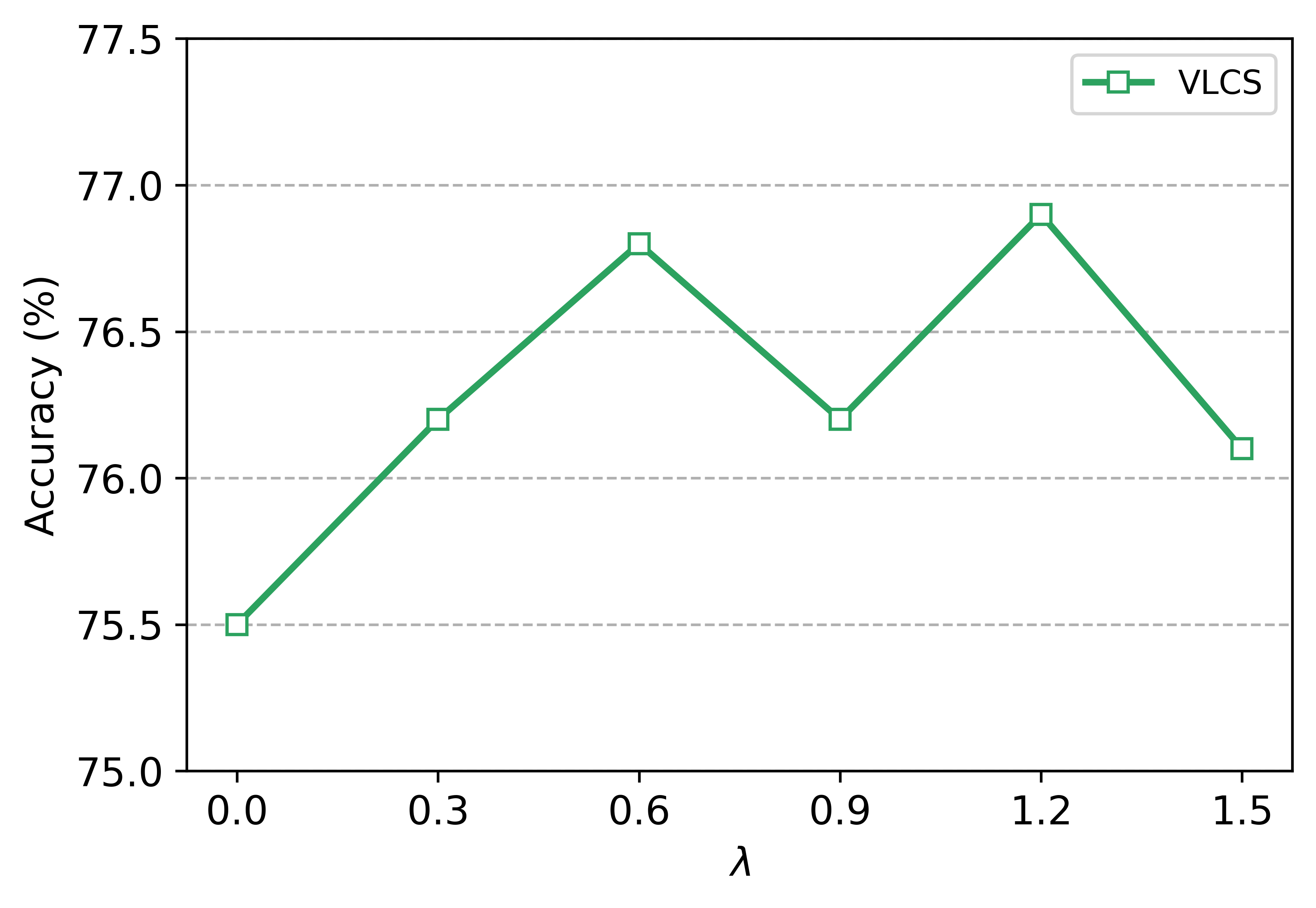}
		\end{minipage}
		\label{fig:lambda_VLCS}
	}
	\subfloat[PACS]{
		\begin{minipage}[b]{0.24\textwidth}
  	 	\includegraphics[width=0.9\textwidth]{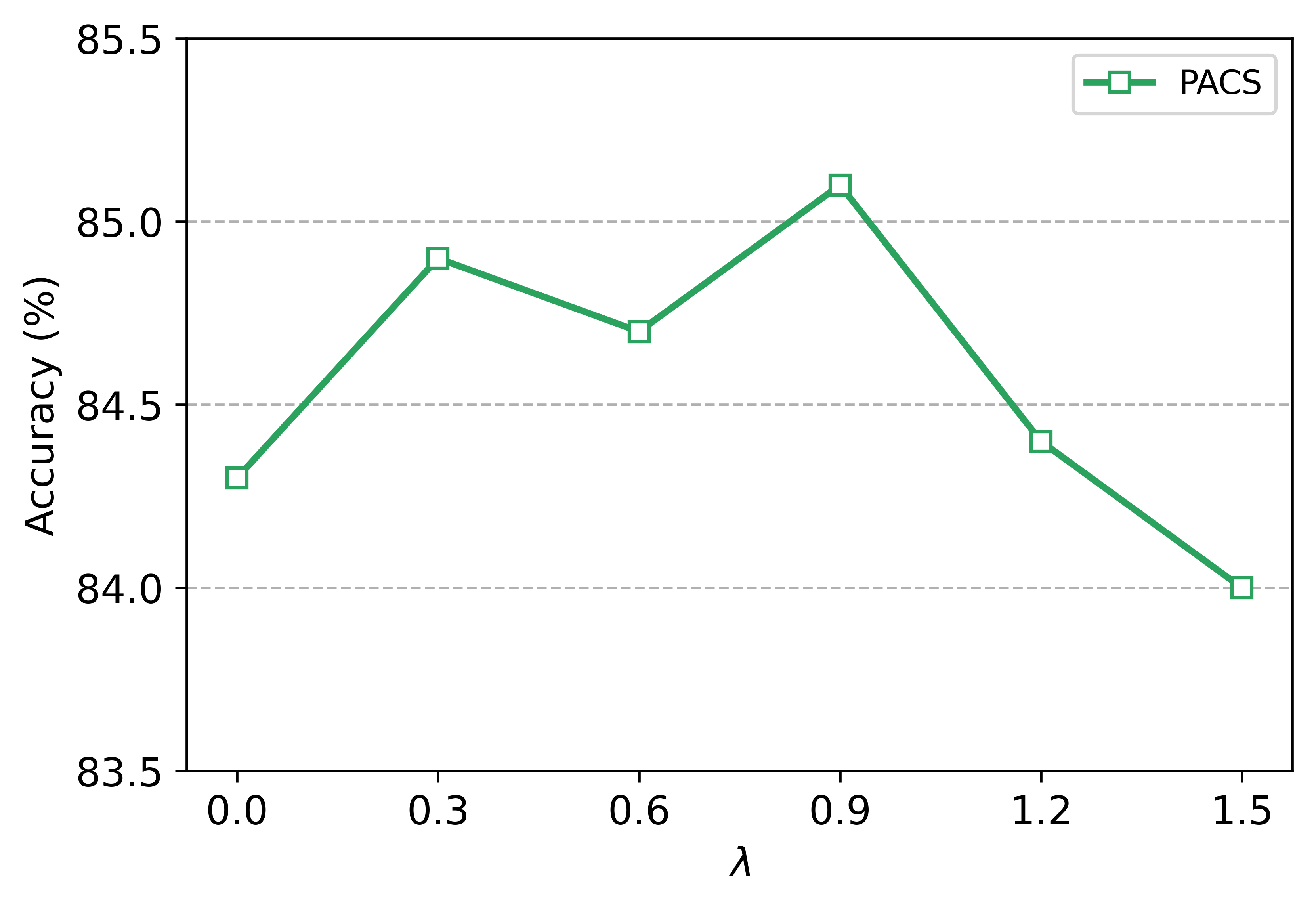}
		\end{minipage}
	\label{fig:lambda_PACS}
	}
        \quad
	\subfloat[OfficeHome]{
		\begin{minipage}[b]{0.24\textwidth}
  	 	\includegraphics[width=0.9\textwidth]{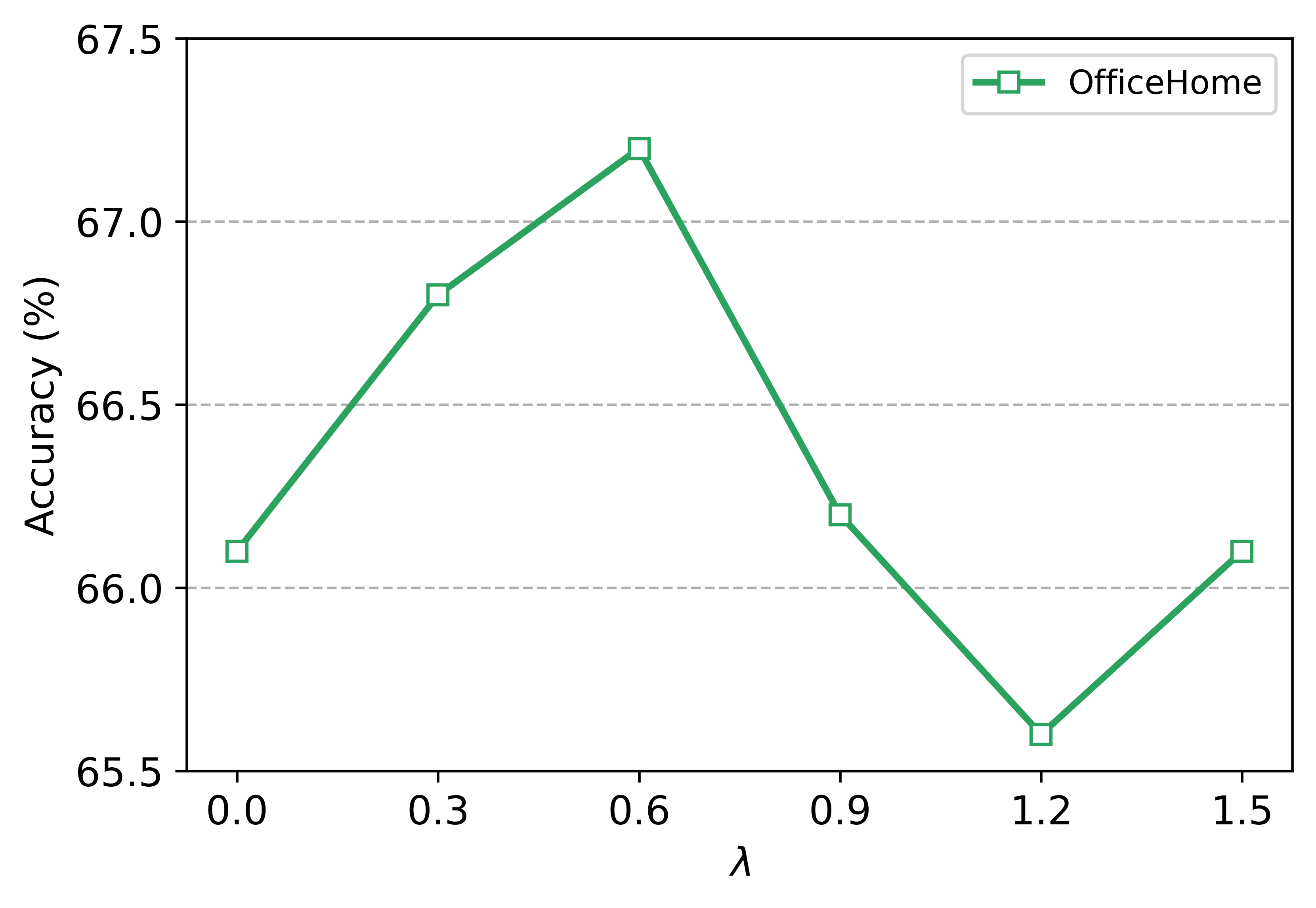}
		\end{minipage}
	\label{fig:lambda_OH}
	}
 	\subfloat[TerraIncognita]{
		\begin{minipage}[b]{0.24\textwidth}
  	 	\includegraphics[width=0.9\textwidth]{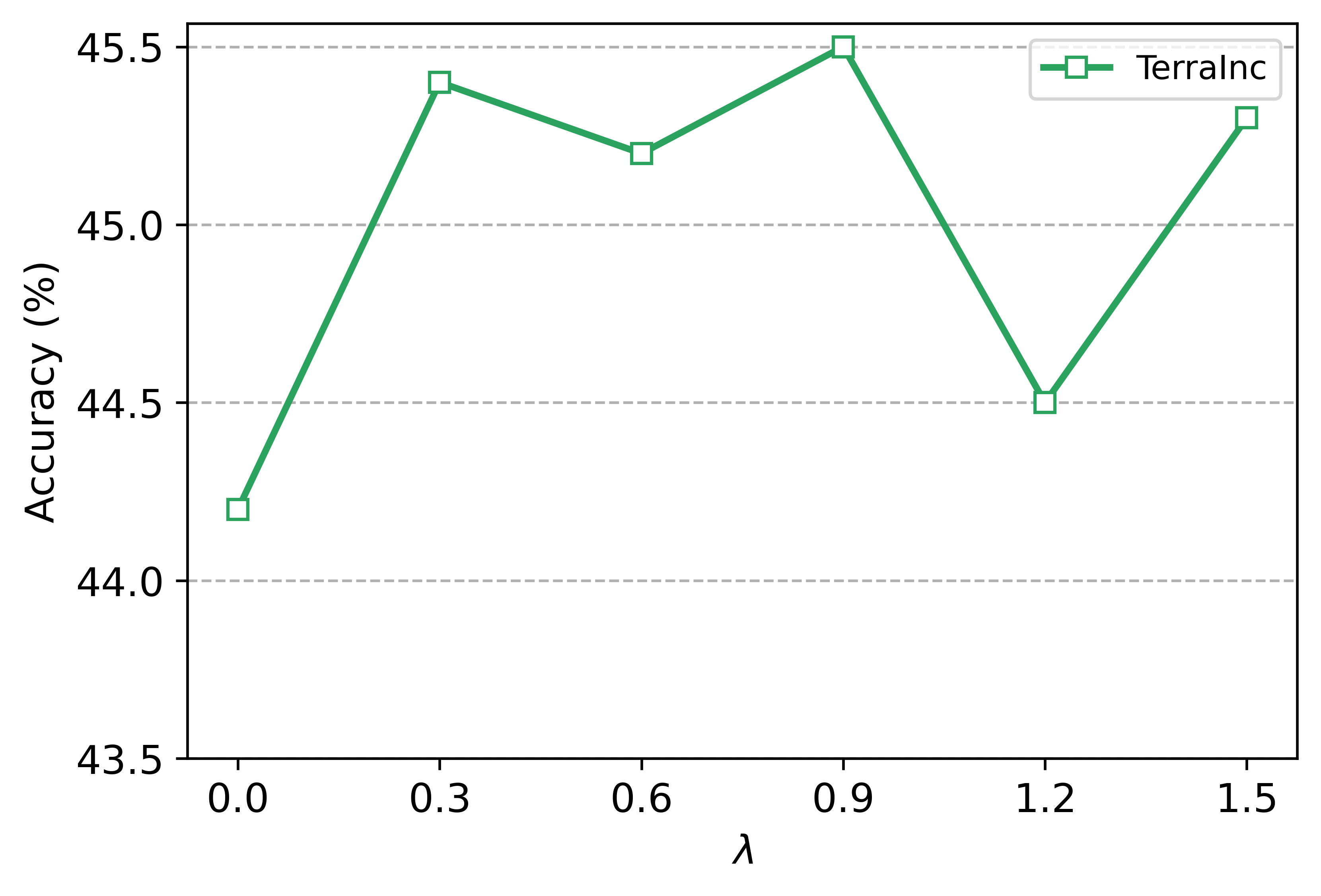}
		\end{minipage}
	\label{fig:lambda_Terra}
	}
\caption{Effect of the loss weight term $\lambda$ for minimizing prediction disagreement score. Experiments are conducted on ERM-based AdaODM. $\lambda$ is set to $\{0, 0.3, 0.6, 0.9, 1.2, 1.5\}$. The other hyperparameters are set by default.}
\label{fig:loss_term}
\end{figure}

\subsection{Ablation Studies}
We further verify the effectiveness of the adversarial training. To do so, we mainly implement two variants of our methods. (1) \textbf{w/o domain-specific classifiers}: we replace domain-specific classifiers with regular classifiers, i.e. each classifier is trained on all source domains. Thus classifiers will not measure domain differences effectively. Adversarial training is also not considered for this setting. (2) \textbf{w/o the prediction disagreement-based adversarial training}: each classifier still receives information from one separate domain but adversarial training is not applied. We conduct the two kinds of variants on all datasets with both ERM and CORAL. Extensive experiments are performed for each target domain ($20 \ trials \ \times \ 3 \ rounds$) in this section. Table \ref{tab2} compares the two variants with the full version of our method. Note that all numbers are performances after conducting test-time online disagreement minimization.

We can make the following observations. (1) For both ERM- and CORAL-based AdaODM, performances drop when domain-specific classifiers or adversarial training strategy is removed. Although classifiers are initialized with different weights, they show similar behavior when receiving the same knowledge instead of domain-specific one. Thus it is difficult for the adversarial training to capture the prediction inconsistency among classifiers. Likewise, without adversarial training, each domain-specific classifier is updated only based on one source domain. The amount of training samples is too small and thus they are more likely to overfit to the corresponding source domains. Minimizing prediction disagreement for the target samples will be less reliable. Compared with the two variants, the full version of our method improves target performance consistently and significantly. This proves the necessity to maximize the discrepancy of source classifiers and generalize domain-specific classifiers to unseen source domains. (2) Comparing the performance gaps between the full AdaODM and the two variants, ERM-based AdaODM is more influenced with either module removed. Since CORAL encourages feature alignment, domain differences in ERM-based experiments are much larger than those in CORAL-based experiments. This again emphasizes the critical roles of our adversarial training strategy for handling large distribution shifts. 

Since we set the test-time batch size to 64 for both AdaODM and baselines. We are also interested in how the performance will be when test samples come one by one, i.e., the test-time batch size equals 1. We evaluate this on ERM-based AdaODM. Results shown in Table \ref{onebyone} indicate that when test samples arrive one by
one, the test results become slightly lower and more fluctuated. This is reasonable
since the gradient direction is less stable when the batch size is small.

\begin{table}
    \caption{Ablation studies to show the effect of test-time batch size on the performance. \textbf{Bold} numbers indicate the best results in each setting.}
    \centering
    \adjustbox{max width=\linewidth }{%
    \begin{tabular}{cccccc}
    \toprule
    \textbf{Test-time batch size} & \textbf{VLCS} & \textbf{PACS} & \textbf{OfficeHome} & \textbf{TerraInc} & \textbf{Avg} \\
    \midrule
    1 & 79.2 $\pm$ 1.3 & 86.8 $\pm$ 1.1 & 68.3 $\pm$ 0.9 & 47.1 $\pm$ 1.6 & 70.5 \\
    64 (default) & \textbf{79.7 $\pm$ 0.8} & \textbf{87.8 $\pm$ 0.5} & \textbf{68.8 $\pm$ 0.2} & \textbf{47.7 $\pm$ 1.2} & \textbf{71.0} \\
    \bottomrule
    \end{tabular}}
    \label{onebyone}
\end{table}

We provide an additional hyper-parameter sensitivity study for the disagreement score $\lambda$. The experiments are conducted on ERM-based AdaODM with default hyperparameters except that the loss weight term $\lambda$ is set in the range of \{0, 0.3, 0.6, 0.9, 1.2, 1.5\}. Fig. \ref{fig:loss_term} illustrates the performance change with different $\lambda$. We can observe that the prediction disagreement minimization loss consistently improves the generalization performance.

\section{Conclusion}
We have proposed an adaptive domain generalization approach AdaODM, which aims to adapt a source model to new domains at test time. The main idea of AdaODM is to: (1) build domain-invariant feature space during training and (2) align target domain features to the invariant feature region at test time, both through promoting prediction consistency. To achieve this, AdaODM learns a domain-generic feature extractor and multiple domain-specific classifiers via a novel prediction disagreement-based adversarial training strategy. Extensive experimental results on four public benchmarks demonstrate that AdaODM is able to align target features effectively, which leads to state-of-the-art performance for domain generalization.


\bibliographystyle{IEEEtran}
\bibliography{IEEEabrv,egbib}

\begin{thebibliography}{10}
\providecommand{\url}[1]{#1}
\csname url@samestyle\endcsname
\providecommand{\newblock}{\relax}
\providecommand{\bibinfo}[2]{#2}
\providecommand{\BIBentrySTDinterwordspacing}{\spaceskip=0pt\relax}
\providecommand{\BIBentryALTinterwordstretchfactor}{4}
\providecommand{\BIBentryALTinterwordspacing}{\spaceskip=\fontdimen2\font plus
\BIBentryALTinterwordstretchfactor\fontdimen3\font minus
  \fontdimen4\font\relax}
\providecommand{\BIBforeignlanguage}[2]{{%
\expandafter\ifx\csname l@#1\endcsname\relax
\typeout{** WARNING: IEEEtran.bst: No hyphenation pattern has been}%
\typeout{** loaded for the language `#1'. Using the pattern for}%
\typeout{** the default language instead.}%
\else
\language=\csname l@#1\endcsname
\fi
#2}}
\providecommand{\BIBdecl}{\relax}
\BIBdecl

\bibitem{gulrajani2020search}
I.~Gulrajani and D.~Lopez-Paz, ``In search of lost domain generalization,'' in
  \emph{International Conference on Learning Representations}, 2020.

\bibitem{vapnik1999overview}
V.~N. Vapnik, ``An overview of statistical learning theory,'' \emph{IEEE
  transactions on neural networks}, vol.~10, no.~5, pp. 988--999, 1999.

\bibitem{wilson2020survey}
G.~Wilson and D.~J. Cook, ``A survey of unsupervised deep domain adaptation,''
  \emph{ACM Transactions on Intelligent Systems and Technology (TIST)},
  vol.~11, no.~5, pp. 1--46, 2020.

\bibitem{csurka2017domain}
G.~Csurka, ``Domain adaptation for visual applications: A comprehensive
  survey,'' \emph{arXiv preprint arXiv:1702.05374}, 2017.

\bibitem{peng2019moment}
X.~Peng, Q.~Bai, X.~Xia, Z.~Huang, K.~Saenko, and B.~Wang, ``Moment matching
  for multi-source domain adaptation,'' in \emph{Proceedings of the IEEE/CVF
  international conference on computer vision}, 2019, pp. 1406--1415.

\bibitem{ganin2015unsupervised}
Y.~Ganin and V.~Lempitsky, ``Unsupervised domain adaptation by
  backpropagation,'' in \emph{International conference on machine
  learning}.\hskip 1em plus 0.5em minus 0.4em\relax PMLR, 2015, pp. 1180--1189.

\bibitem{tzeng2017adversarial}
E.~Tzeng, J.~Hoffman, K.~Saenko, and T.~Darrell, ``Adversarial discriminative
  domain adaptation,'' in \emph{Proceedings of the IEEE conference on computer
  vision and pattern recognition}, 2017, pp. 7167--7176.

\bibitem{hu2021mixnorm}
X.~Hu, G.~Uzunbas, S.~Chen, R.~Wang, A.~Shah, R.~Nevatia, and S.-N. Lim,
  ``Mixnorm: Test-time adaptation through online normalization estimation,''
  \emph{arXiv preprint arXiv:2110.11478}, 2021.

\bibitem{schneider2020removing}
S.~Schneider, E.~Rusak, L.~Eck, O.~Bringmann, W.~Brendel, and M.~Bethge,
  ``Removing covariate shift improves robustness against common corruptions,''
  in \emph{Thirty-fourth Conference on Neural Information Processing Systems
  (NeurIPS)}, 2020.

\bibitem{liang2020we}
J.~Liang, D.~Hu, and J.~Feng, ``Do we really need to access the source data?
  source hypothesis transfer for unsupervised domain adaptation,'' in
  \emph{International Conference on Machine Learning}.\hskip 1em plus 0.5em
  minus 0.4em\relax PMLR, 2020, pp. 6028--6039.

\bibitem{wang2020tent}
D.~Wang, E.~Shelhamer, S.~Liu, B.~Olshausen, and T.~Darrell, ``Tent: Fully
  test-time adaptation by entropy minimization,'' in \emph{International
  Conference on Learning Representations}, 2020.

\bibitem{iwasawa2021test}
Y.~Iwasawa and Y.~Matsuo, ``Test-time classifier adjustment module for
  model-agnostic domain generalization,'' \emph{Advances in Neural Information
  Processing Systems}, vol.~34, 2021.

\bibitem{luo2008transfer}
P.~Luo, F.~Zhuang, H.~Xiong, Y.~Xiong, and Q.~He, ``Transfer learning from
  multiple source domains via consensus regularization,'' in \emph{Proceedings
  of the 17th ACM conference on Information and knowledge management}, 2008,
  pp. 103--112.

\bibitem{long2017deep}
M.~Long, H.~Zhu, J.~Wang, and M.~I. Jordan, ``Deep transfer learning with joint
  adaptation networks,'' in \emph{International conference on machine
  learning}.\hskip 1em plus 0.5em minus 0.4em\relax PMLR, 2017, pp. 2208--2217.

\bibitem{gretton2012kernel}
A.~Gretton, K.~M. Borgwardt, M.~J. Rasch, B.~Sch{\"o}lkopf, and A.~Smola, ``A
  kernel two-sample test,'' \emph{The Journal of Machine Learning Research},
  vol.~13, no.~1, pp. 723--773, 2012.

\bibitem{sun2016deep}
B.~Sun and K.~Saenko, ``Deep coral: Correlation alignment for deep domain
  adaptation,'' in \emph{European conference on computer vision}.\hskip 1em
  plus 0.5em minus 0.4em\relax Springer, 2016, pp. 443--450.

\bibitem{ganin2016domain}
Y.~Ganin, E.~Ustinova, H.~Ajakan, P.~Germain, H.~Larochelle, F.~Laviolette,
  M.~Marchand, and V.~Lempitsky, ``Domain-adversarial training of neural
  networks,'' \emph{The journal of machine learning research}, vol.~17, no.~1,
  pp. 2096--2030, 2016.

\bibitem{saito2018maximum}
K.~Saito, K.~Watanabe, Y.~Ushiku, and T.~Harada, ``Maximum classifier
  discrepancy for unsupervised domain adaptation,'' in \emph{Proceedings of the
  IEEE conference on computer vision and pattern recognition}, 2018, pp.
  3723--3732.

\bibitem{long2018conditional}
M.~Long, Z.~Cao, J.~Wang, and M.~I. Jordan, ``Conditional adversarial domain
  adaptation,'' \emph{Advances in neural information processing systems},
  vol.~31, 2018.

\bibitem{hoffman2018cycada}
J.~Hoffman, E.~Tzeng, T.~Park, J.-Y. Zhu, P.~Isola, K.~Saenko, A.~Efros, and
  T.~Darrell, ``Cycada: Cycle-consistent adversarial domain adaptation,'' in
  \emph{International conference on machine learning}.\hskip 1em plus 0.5em
  minus 0.4em\relax PMLR, 2018, pp. 1989--1998.

\bibitem{li2020model}
R.~Li, Q.~Jiao, W.~Cao, H.-S. Wong, and S.~Wu, ``Model adaptation: Unsupervised
  domain adaptation without source data,'' in \emph{Proceedings of the IEEE/CVF
  Conference on Computer Vision and Pattern Recognition}, 2020, pp. 9641--9650.

\bibitem{kundu2020universal}
J.~N. Kundu, N.~Venkat, R.~V. Babu \emph{et~al.}, ``Universal source-free
  domain adaptation,'' in \emph{Proceedings of the IEEE/CVF Conference on
  Computer Vision and Pattern Recognition}, 2020, pp. 4544--4553.

\bibitem{li2018deep}
Y.~Li, X.~Tian, M.~Gong, Y.~Liu, T.~Liu, K.~Zhang, and D.~Tao, ``Deep domain
  generalization via conditional invariant adversarial networks,'' in
  \emph{Proceedings of the European Conference on Computer Vision (ECCV)},
  2018, pp. 624--639.

\bibitem{li2018domain}
H.~Li, S.~J. Pan, S.~Wang, and A.~C. Kot, ``Domain generalization with
  adversarial feature learning,'' in \emph{Proceedings of the IEEE conference
  on computer vision and pattern recognition}, 2018, pp. 5400--5409.

\bibitem{senerdomain}
O.~Sener and V.~Koltun, ``Domain generalization without excess empirical
  risk,'' in \emph{Advances in Neural Information Processing Systems}.

\bibitem{dingdomain}
Y.~Ding, L.~Wang, B.~Liang, S.~Liang, Y.~Wang, and F.~Chen, ``Domain
  generalization by learning and removing domain-specific features,'' in
  \emph{Advances in Neural Information Processing Systems}.

\bibitem{li2018learning}
D.~Li, Y.~Yang, Y.-Z. Song, and T.~M. Hospedales, ``Learning to generalize:
  Meta-learning for domain generalization,'' in \emph{Thirty-Second AAAI
  Conference on Artificial Intelligence}, 2018.

\bibitem{dou2019domain}
Q.~Dou, D.~Coelho~de Castro, K.~Kamnitsas, and B.~Glocker, ``Domain
  generalization via model-agnostic learning of semantic features,''
  \emph{Advances in Neural Information Processing Systems}, vol.~32, 2019.

\bibitem{balaji2018metareg}
Y.~Balaji, S.~Sankaranarayanan, and R.~Chellappa, ``Metareg: Towards domain
  generalization using meta-regularization,'' \emph{Advances in neural
  information processing systems}, vol.~31, 2018.

\bibitem{arjovsky2019invariant}
M.~Arjovsky, L.~Bottou, I.~Gulrajani, and D.~Lopez-Paz, ``Invariant risk
  minimization,'' \emph{arXiv preprint arXiv:1907.02893}, 2019.

\bibitem{wang2020heterogeneous}
Y.~Wang, H.~Li, and A.~C. Kot, ``Heterogeneous domain generalization via domain
  mixup,'' in \emph{ICASSP 2020-2020 IEEE International Conference on
  Acoustics, Speech and Signal Processing (ICASSP)}.\hskip 1em plus 0.5em minus
  0.4em\relax IEEE, 2020, pp. 3622--3626.

\bibitem{xu2020adversarial}
M.~Xu, J.~Zhang, B.~Ni, T.~Li, C.~Wang, Q.~Tian, and W.~Zhang, ``Adversarial
  domain adaptation with domain mixup,'' in \emph{Proceedings of the AAAI
  Conference on Artificial Intelligence}, vol.~34, no.~04, 2020, pp.
  6502--6509.

\bibitem{yan2020improve}
S.~Yan, H.~Song, N.~Li, L.~Zou, and L.~Ren, ``Improve unsupervised domain
  adaptation with mixup training,'' \emph{arXiv preprint arXiv:2001.00677},
  2020.

\bibitem{zhou2020domain}
K.~Zhou, Y.~Yang, Y.~Qiao, and T.~Xiang, ``Domain generalization with
  mixstyle,'' in \emph{International Conference on Learning Representations},
  2020.

\bibitem{jeon2021feature}
S.~Jeon, K.~Hong, P.~Lee, J.~Lee, and H.~Byun, ``Feature stylization and
  domain-aware contrastive learning for domain generalization,'' in
  \emph{Proceedings of the 29th ACM International Conference on Multimedia},
  2021, pp. 22--31.

\bibitem{wang2021feature}
Y.~Wang, L.~Qi, Y.~Shi, and Y.~Gao, ``Feature-based style randomization for
  domain generalization,'' \emph{arXiv preprint arXiv:2106.03171}, 2021.

\bibitem{sun2020test}
Y.~Sun, X.~Wang, Z.~Liu, J.~Miller, A.~Efros, and M.~Hardt, ``Test-time
  training with self-supervision for generalization under distribution
  shifts,'' in \emph{International Conference on Machine Learning}.\hskip 1em
  plus 0.5em minus 0.4em\relax PMLR, 2020, pp. 9229--9248.

\bibitem{yang2022domain}
S.~Yang, D.~Das, J.~Cho, H.~Park, and S.~Yun, ``Domain agnostic few-shot
  learning for speaker verification,'' \emph{arXiv preprint arXiv:2206.13700},
  2022.

\bibitem{liu2021ttt++}
Y.~Liu, P.~Kothari, B.~van Delft, B.~Bellot-Gurlet, T.~Mordan, and A.~Alahi,
  ``Ttt++: When does self-supervised test-time training fail or thrive?''
  \emph{Advances in Neural Information Processing Systems}, vol.~34, 2021.

\bibitem{lee2013pseudo}
D.-H. Lee \emph{et~al.}, ``Pseudo-label: The simple and efficient
  semi-supervised learning method for deep neural networks,'' in \emph{Workshop
  on challenges in representation learning, ICML}, vol.~3, no.~2, 2013, p. 896.

\bibitem{zhang2021memo}
M.~Zhang, S.~Levine, and C.~Finn, ``Memo: Test time robustness via adaptation
  and augmentation,'' \emph{arXiv preprint arXiv:2110.09506}, 2021.

\bibitem{miyato2018virtual}
T.~Miyato, S.-i. Maeda, M.~Koyama, and S.~Ishii, ``Virtual adversarial
  training: a regularization method for supervised and semi-supervised
  learning,'' \emph{IEEE transactions on pattern analysis and machine
  intelligence}, vol.~41, no.~8, pp. 1979--1993, 2018.

\bibitem{xie2020unsupervised}
Q.~Xie, Z.~Dai, E.~Hovy, T.~Luong, and Q.~Le, ``Unsupervised data augmentation
  for consistency training,'' \emph{Advances in Neural Information Processing
  Systems}, vol.~33, pp. 6256--6268, 2020.

\bibitem{zhang2019consistency}
H.~Zhang, Z.~Zhang, A.~Odena, and H.~Lee, ``Consistency regularization for
  generative adversarial networks,'' in \emph{International Conference on
  Learning Representations}, 2019.

\bibitem{sinha2021consistency}
S.~Sinha and A.~B. Dieng, ``Consistency regularization for variational
  auto-encoders,'' \emph{Advances in Neural Information Processing Systems},
  vol.~34, 2021.

\bibitem{tarvainen2017mean}
A.~Tarvainen and H.~Valpola, ``Mean teachers are better role models:
  Weight-averaged consistency targets improve semi-supervised deep learning
  results,'' \emph{Advances in neural information processing systems}, vol.~30,
  2017.

\bibitem{french2017self}
G.~French, M.~Mackiewicz, and M.~Fisher, ``Self-ensembling for visual domain
  adaptation,'' \emph{arXiv preprint arXiv:1706.05208}, 2017.

\bibitem{wu2020dual}
Y.~Wu, D.~Inkpen, and A.~El-Roby, ``Dual mixup regularized learning for
  adversarial domain adaptation,'' in \emph{European Conference on Computer
  Vision}.\hskip 1em plus 0.5em minus 0.4em\relax Springer, 2020, pp. 540--555.

\bibitem{xu2021fourier}
Q.~Xu, R.~Zhang, Y.~Zhang, Y.~Wang, and Q.~Tian, ``A fourier-based framework
  for domain generalization,'' in \emph{Proceedings of the IEEE/CVF Conference
  on Computer Vision and Pattern Recognition}, 2021, pp. 14\,383--14\,392.

\bibitem{vanschoren2019meta}
J.~Vanschoren, ``Meta-learning,'' in \emph{Automated Machine Learning}.\hskip
  1em plus 0.5em minus 0.4em\relax Springer, Cham, 2019, pp. 35--61.

\bibitem{fang2013unbiased}
C.~Fang, Y.~Xu, and D.~N. Rockmore, ``Unbiased metric learning: On the
  utilization of multiple datasets and web images for softening bias,'' in
  \emph{Proceedings of the IEEE International Conference on Computer Vision},
  2013, pp. 1657--1664.

\bibitem{li2017deeper}
D.~Li, Y.~Yang, Y.-Z. Song, and T.~M. Hospedales, ``Deeper, broader and artier
  domain generalization,'' in \emph{Proceedings of the IEEE international
  conference on computer vision}, 2017, pp. 5542--5550.

\bibitem{venkateswara2017deep}
H.~Venkateswara, J.~Eusebio, S.~Chakraborty, and S.~Panchanathan, ``Deep
  hashing network for unsupervised domain adaptation,'' in \emph{Proceedings of
  the IEEE conference on computer vision and pattern recognition}, 2017, pp.
  5018--5027.

\bibitem{beery2018recognition}
S.~Beery, G.~Van~Horn, and P.~Perona, ``Recognition in terra incognita,'' in
  \emph{Proceedings of the European conference on computer vision (ECCV)},
  2018, pp. 456--473.

\bibitem{sagawa2019distributionally}
S.~Sagawa, P.~W. Koh, T.~B. Hashimoto, and P.~Liang, ``Distributionally robust
  neural networks for group shifts: On the importance of regularization for
  worst-case generalization,'' \emph{arXiv preprint arXiv:1911.08731}, 2019.

\bibitem{blanchard2011generalizing}
G.~Blanchard, G.~Lee, and C.~Scott, ``Generalizing from several related
  classification tasks to a new unlabeled sample,'' \emph{Advances in neural
  information processing systems}, vol.~24, 2011.

\bibitem{nam2019reducing}
H.~Nam, H.~Lee, J.~Park, W.~Yoon, and D.~Yoo, ``Reducing domain gap via
  style-agnostic networks,'' \emph{arXiv preprint arXiv:1910.11645}, vol.~2,
  no.~7, p.~8, 2019.

\bibitem{zhang2021adaptive}
M.~Zhang, H.~Marklund, N.~Dhawan, A.~Gupta, S.~Levine, and C.~Finn, ``Adaptive
  risk minimization: Learning to adapt to domain shift,'' \emph{Advances in
  Neural Information Processing Systems}, vol.~34, 2021.

\bibitem{krueger2021out}
D.~Krueger, E.~Caballero, J.-H. Jacobsen, A.~Zhang, J.~Binas, D.~Zhang,
  R.~Le~Priol, and A.~Courville, ``Out-of-distribution generalization via risk
  extrapolation (rex),'' in \emph{International Conference on Machine
  Learning}.\hskip 1em plus 0.5em minus 0.4em\relax PMLR, 2021, pp. 5815--5826.

\bibitem{huang2020self}
Z.~Huang, H.~Wang, E.~P. Xing, and D.~Huang, ``Self-challenging improves
  cross-domain generalization,'' in \emph{European Conference on Computer
  Vision}.\hskip 1em plus 0.5em minus 0.4em\relax Springer, 2020, pp. 124--140.

\bibitem{bui2021exploiting}
M.-H. Bui, T.~Tran, A.~Tran, and D.~Phung, ``Exploiting domain-specific
  features to enhance domain generalization,'' \emph{Advances in Neural
  Information Processing Systems}, vol.~34, 2021.

\bibitem{dubey2021adaptive}
A.~Dubey, V.~Ramanathan, A.~Pentland, and D.~Mahajan, ``Adaptive methods for
  real-world domain generalization,'' in \emph{Proceedings of the IEEE/CVF
  Conference on Computer Vision and Pattern Recognition}, 2021, pp.
  14\,340--14\,349.

\bibitem{van2008visualizing}
L.~Van~der Maaten and G.~Hinton, ``Visualizing data using t-sne.''
  \emph{Journal of machine learning research}, vol.~9, no.~11, 2008.

\end{thebibliography}

\end{document}